\documentclass{article}

\usepackage[preprint]{colm2026_conference}

\usepackage[utf8]{inputenc}
\usepackage[T1]{fontenc}
\usepackage{hyperref}
\usepackage{url}
\usepackage{booktabs}
\usepackage{amsfonts}
\usepackage{amsmath}
\usepackage{amssymb}
\usepackage{nicefrac}
\usepackage{microtype}
\usepackage{graphicx}
\usepackage[table]{xcolor}
\usepackage{subcaption}
\usepackage{algorithm}
\usepackage{algorithmic}
\usepackage{multirow}
\usepackage{wrapfig}
\usepackage{colortbl}
\usepackage{array}
\usepackage{tcolorbox}
\tcbuselibrary{breakable}
\usepackage{pifont}
\usepackage{arydshln}
\usepackage{pgfplots}
\pgfplotsset{compat=1.18}
\usepgfplotslibrary{groupplots}
\usepackage{enumitem}
\usepackage{subcaption}
\usepackage{placeins}   % provides \FloatBarrier

\definecolor{darkblue}{rgb}{0, 0, 0.5}
\definecolor{lightgray}{gray}{0.92}
\definecolor{successgreen}{rgb}{0.0, 0.5, 0.0}
\definecolor{failred}{rgb}{0.7, 0.0, 0.0}
\hypersetup{colorlinks=true, citecolor=darkblue, linkcolor=darkblue, urlcolor=darkblue}

\newcommand{\ours}{\textsc{Omni-SimpleMem}}
\newcommand{\tool}{\textsc{AutoResearchClaw}}

\newcommand{\eg}{\textit{e.g.}}

\newcommand{\xmark}{\ding{55}}

\title{\textsc{Omni-SimpleMem}: Autoresearch-Guided Discovery of Lifelong Multimodal Agent Memory}

\author{
    Jiaqi Liu$^{1}$, Zipeng Ling$^2$, Shi Qiu $^1$,  Yanqing Liu$^3$,  Siwei Han$^{1}$, Peng Xia$^{1}$, Haoqin Tu$^{3}$, \\ 
    \textbf{Zeyu Zheng}$^{4}$, \textbf{Cihang Xie}$^{3}$,  \textbf{Charles Fleming}$^{5}$, \textbf{Mingyu Ding}$^{1}$, \textbf{Huaxiu Yao}$^{1}$
    \\
    $^1$UNC-Chapel Hill\quad
    $^2$University of Pennsylvania \quad
    $^3$University of California, Santa Cruz\quad \\
    $^4$University of California Berkeley\quad
    $^5$Cisco
    \\
 \texttt{
    jqliu@cs.unc.edu \quad
    huaxiu@cs.unc.edu
    }
}
\begin{document}

\maketitle

\vspace{-1em}
\begin{abstract}
\vspace{-0.5em}
AI agents increasingly operate over extended time horizons, yet their ability to retain, organize, and recall multimodal experiences remains a critical bottleneck. 
Building effective lifelong memory requires navigating a vast design space spanning architecture, retrieval strategies, prompt engineering, and data pipelines; this space is too large and interconnected for manual exploration or traditional AutoML to explore effectively.
We deploy an autonomous research pipeline to discover \ours{}, a unified multimodal memory framework for lifelong AI agents. Starting from a na\"ive baseline (F1\,=\,0.117 on LoCoMo), the pipeline autonomously executes ${\sim}50$ experiments across two benchmarks, diagnosing failure modes, proposing architectural modifications, and repairing data pipeline bugs, all without human intervention in the inner loop. The resulting system achieves state-of-the-art on both benchmarks, improving F1 by +411\% on LoCoMo and +214\% on Mem-Gallery relative to the initial configurations. Critically, the most impactful discoveries are not hyperparameter adjustments: bug fixes (+175\%), architectural changes (+44\%), and prompt engineering (+188\% on specific categories) each individually exceed the cumulative contribution of all hyperparameter tuning, demonstrating capabilities fundamentally beyond the reach of traditional AutoML. We provide a taxonomy of six discovery types and identify four properties that make multimodal memory particularly suited for autoresearch, offering guidance for applying autonomous research pipelines to other AI system domains. Code is available at
this \href{https://github.com/aiming-lab/SimpleMem}{https://github.com/aiming-lab/SimpleMem}.
\end{abstract}

\begin{figure*}[h]
\vspace{-1.5em}
\centering
\includegraphics[width=0.95\textwidth]{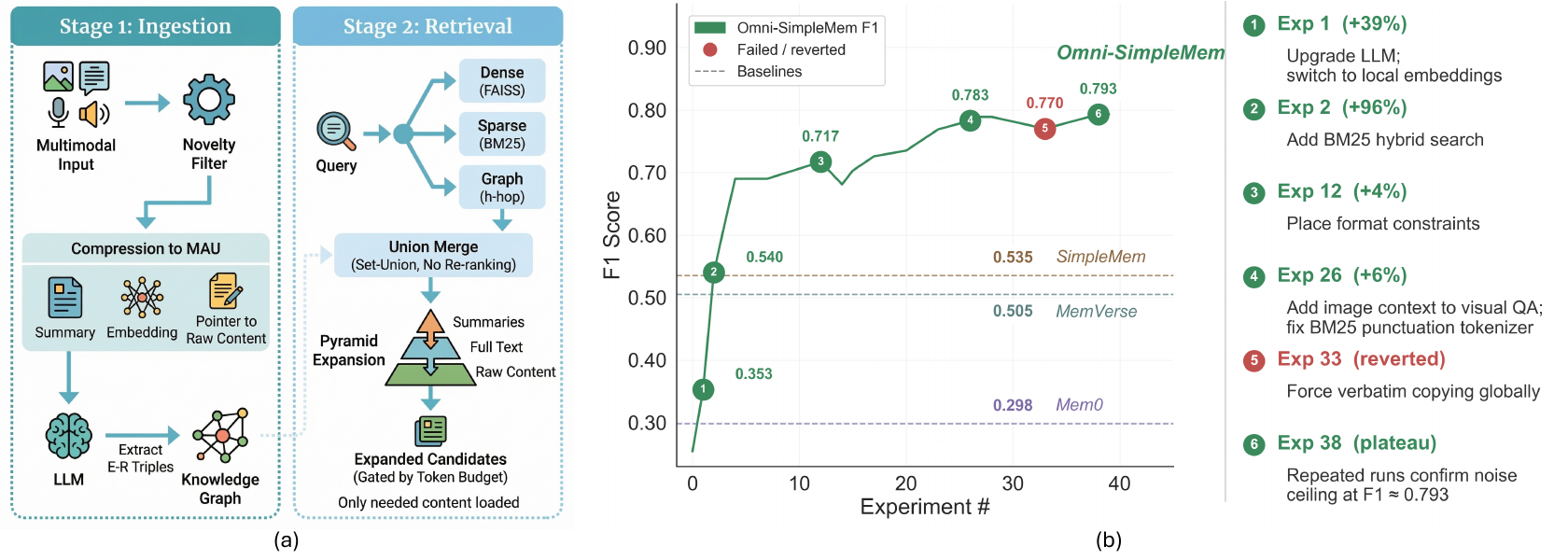}
\caption{\textbf{Overview of the Discovery Process of \ours{}.} \textbf{(a)}~Discovered architecture: multimodal inputs are filtered for novelty, compressed into MAUs, and retrieved via hybrid dense-sparse-graph search with pyramid expansion. \textbf{(b)}~Autonomous optimization trajectory on Mem-Gallery: 39 experiments improve F1 from 0.254 to 0.793 (+214\%).}
\label{fig:overview}
\end{figure*} 
\vspace{-0.5em}
\section{Introduction}
\vspace{-0.5em}
\label{sec:intro}

Recent advances in large language models have given rise to AI agents capable of tool use, multi-step reasoning, and cross-modal comprehension~\citep{yao2022react,yan2025positionmultimodallargelanguage}. These agents interact with users over extended time horizons, accumulating diverse streams of text, images, audio, and video throughout their operation. However, their ability to retain, organize, and recall past experiences remains a critical bottleneck~\citep{zhang2024surveymemorymechanismlarge,xu2025amemagenticmemoryllm}. Building effective lifelong multimodal memory requires navigating a vast design space spanning architectural choices (how to structure storage), retrieval strategies (how to find relevant information), prompt engineering (how to present context to the LLM), and data pipeline configurations (how to ingest and process heterogeneous inputs).

Existing approaches to agent memory fall into two broad categories, each with notable limitations. The first stores raw inputs and retrieves them via embedding similarity~\citep{lewis2020rag,borgeaud2022retro}, suffering from storage bloat and retrieval noise as the memory grows. The second introduces structured memory management with explicit operations~\citep{packer2023memgpt,park2023generative}, but typically operates on text alone, discarding rich visual and auditory signals. Crucially, both categories are products of manual research cycles: a human researcher hypothesizes an improvement, implements it, evaluates on a benchmark, and iterates. A single researcher may explore only a handful of configurations per day, and important interactions between tightly coupled components are easily missed. Traditional AutoML methods~\citep{hutter2019automated} can search over predefined numerical hyperparameter spaces, but cannot perform the code comprehension, bug diagnosis, architectural redesign, and cross-component reasoning that account for the largest performance gains in complex systems. As a result, existing memory systems inherit the blind spots of their designers--limitations that a more systematic search could avoid.

Recent work on autonomous scientific discovery~\citep{lu2024aiscientist,romeraparedes2024funsearch,panfilov2026claudini} has shown that LLM agents can autonomously discover novel algorithms that outperform human-designed baselines, provided the target domain admits well-defined, quantitative evaluation signals. We ask whether this paradigm extends to complex, multi-component AI systems and answer affirmatively. We deploy \tool{}~\citep{liu2026autoresearchclaw}, a 23-stage autonomous research pipeline, to discover \ours{}, a unified multimodal memory framework for lifelong AI agents. Starting from a na\"ive baseline (F1\,=\,0.117 on LoCoMo), the pipeline autonomously executes ${\sim}50$ experiments across two benchmarks, iteratively diagnosing failure modes, proposing architectural modifications, repairing data pipeline bugs, and validating improvements, all without human intervention in the inner loop. The resulting system achieves state-of-the-art on both benchmarks, improving F1 by +411\% on LoCoMo (0.117$\to$0.598) and +214\% on Mem-Gallery (0.254$\to$0.797) relative to the initial configurations. Critically, the most impactful discoveries are \emph{not} hyperparameter adjustments: bug fixes (+175\%), architectural changes (+44\%), and prompt engineering (+188\% on specific categories) each individually exceed the cumulative contribution of all hyperparameter tuning, demonstrating capabilities fundamentally beyond the reach of traditional AutoML.

Among the pipeline's most consequential discoveries are three architectural principles that define \ours{}. 
% \QS{The representation is inconsistent with Section 3.2: selective ingestion / unified representation / progressive retrieval here, but selective ingestion / progressive retrieval / structured knowledge at chapter 3.2. We need to unify the representations.}
First, \textbf{selective ingestion}: lightweight perceptual encoders measure the information novelty of each incoming signal and discard redundant content before storage, significantly reducing storage requirements. Second, \textbf{unified representation}: all memories, regardless of modality, are represented as Multimodal Atomic Units (MAUs) that separate lightweight metadata from heavy raw data, enabling fast search over compact metadata while preserving full-content access on demand. Third, \textbf{progressive retrieval}: a pyramid mechanism expands information in three stages (summaries, details, raw evidence), each gated by a token budget, backed by a hybrid search strategy combining dense vector retrieval with sparse keyword matching via set-union merging, a strategy autonomously discovered by the pipeline. Our key observation is that multimodal memory is particularly well-suited for autonomous research pipelines due to four properties: immediate scalar evaluation metrics enabling tight optimization loops, modular architecture allowing isolated component modification, fast iteration cycles (1--2 hours per experiment) supporting dozens of hypotheses within days, and version-controlled code modifications allowing failed experiments to be cleanly reverted.

In summary, our primary contribution is \ours{}, a unified multimodal memory framework whose architecture and configuration are discovered through \tool{} and that achieves state-of-the-art results on both evaluated benchmarks.
Beyond the system itself, we provide a comprehensive taxonomy of autonomous discoveries across ${\sim}50$ experiments, revealing that the highest-impact improvements lie beyond the reach of traditional AutoML, and we characterize the convergence behavior, failure modes, and automatic recovery patterns of the pipeline. Our analysis further identifies four properties that make multimodal memory a particularly suitable domain for autoresearch, providing guidance for future applications to other AI system domains. 

% We open-source the complete framework, all benchmark harnesses, and full experiment logs.

\vspace{-1em}
\section{Related Work}
\vspace{-0.5em}
\label{sec:related}

\noindent \textbf{Autonomous Scientific Discovery.}
The vision of AI-driven research has advanced rapidly.
\emph{The AI Scientist} \citep{lu2024aiscientist} demonstrated end-to-end paper generation at ${\sim}\$15$ per paper across three ML domains, with its successor \emph{AI Scientist v2} \citep{lu2025aiscientistv2} eliminating human-authored templates through agentic tree search.
\emph{FunSearch} \citep{romeraparedes2024funsearch} pairs LLM creativity with programmatic evaluation to discover novel mathematical constructions.
\tool{} \citep{liu2026autoresearchclaw} introduces a 23-stage autonomous research pipeline with multi-agent debate and self-healing execution.
\emph{AI-Researcher} \citep{tang2025airesearcher} introduces collaborative multi-agent frameworks with structured iterative refinement, and
\emph{Bilevel Autoresearch} \citep{qu2026bilevel} meta-optimizes the search strategy itself.
A comprehensive survey by \citet{tie2025survey} maps the evolution from foundational modules (2022--2023) through closed-loop systems (2024) to scalable human-AI collaboration (2025+).
We apply the autoresearch paradigm to multi-component AI system optimization, where the challenge shifts from discovering isolated artifacts to diagnosing and improving interactions across tightly coupled modules.

\noindent \textbf{Multimodal Memory Systems.}
Memory-augmented LLM agents have evolved from text-only systems, including MemGPT \citep{packer2023memgpt} with OS-inspired memory hierarchies, Generative Agents \citep{park2023generative} with recency-importance-relevance scoring, SimpleMem \citep{liu2026simplemem} with efficient lifelong memory, and A-Mem \citep{xu2025amem} with LLM-directed reorganization, to multimodal architectures.
MemVerse \citep{memverse2024} combines episodic-semantic memory with multimodal knowledge graphs but requires three LLM calls per ingested item.
Mem0 \citep{mem02024} offers dynamic fact extraction with optional graph memory.
VisRAG \citep{visrag2024} indexes visual pages directly, avoiding text extraction losses.
Claude-Mem \citep{anthropic2024claude} provides commercial embedding-based dialogue memory.
These systems all require extensive manual tuning of retrieval strategies, ingestion pipelines, and prompt configurations, which is precisely the kind of optimization that autonomous research pipelines can accelerate.

\noindent \textbf{Automated Machine Learning.}
Neural Architecture Search~\citep{hutter2019automl,zoph2017nas,liu2019darts} automates model design but operates primarily on well-defined architectural search spaces with differentiable or reinforcement-learning-based objectives.
Hyperparameter optimization methods \citep{bergstra2011tpe,falkner2018bohb} efficiently navigate continuous and categorical spaces, while systems like Auto-sklearn~2.0 \citep{feurer2022autosklearn20handsfreeautoml} automate full ML pipelines including preprocessing and model selection via meta-learning.
More recently, LLM-based agents have been applied to ML tasks: MLAgentBench \citep{huang2024mlagentbenchevaluatinglanguageagents} benchmarks LLM agents on ML research tasks involving code modification, demonstrating the potential of language-guided optimization.
Our setting differs fundamentally: the ``search space'' includes not only hyperparameters and architectural choices, but also \emph{prompt engineering}, \emph{data pipeline bug detection and repair}, \emph{evaluation format alignment}, and \emph{cross-component interaction diagnosis}, all of which require natural language understanding and code modification capabilities beyond traditional AutoML.

\vspace{-0.8em}
\section{Autoresearch-Guided Discovery of \ours{}}
\vspace{-0.8em}
\label{sec:method}

In this section, we describe the autonomous optimization process and the system it produces. We first overview the pipeline (\S\ref{sec:method:pipeline}), then present the discovered \ours{} architecture (\S\ref{sec:method:arch}), followed by the benchmark-specific optimization strategy (\S\ref{sec:method:optimization}).

\subsection{Pipeline Overview}
\label{sec:method:pipeline}

As discussed in Section~\ref{sec:intro}, the design space of multimodal memory systems is too large and interconnected for manual exploration to cover effectively. To address this, we deploy \tool{}~\citep{liu2026autoresearchclaw}, a 23-stage autonomous research pipeline, to systematically optimize \ours{}.
The pipeline receives three inputs: (1)~the SimpleMem~\citep{liu2026simplemem} codebase, a unimodal text-only lifelong memory framework, as a starting point, (2)~two benchmark evaluation harnesses with quantitative metrics (F1), and (3)~API access to LLM providers.
It then enters an iterative loop: at each step, the pipeline analyzes prior results, generates a hypothesis for improvement, implements the change in code, evaluates on a benchmark, and decides whether to \textbf{proceed} (metric improved by $\geq 0.5\%$), \textbf{iterate} (ambiguous result; refine the current hypothesis), or \textbf{pivot} (two consecutive degradations; revert and try a new direction).
Of ${\sim}50$ total experiments, most resulted in a proceed decision, with the remainder split between iterate and pivot.
The full pipeline phases (scoping, literature discovery, multi-agent debate, experiment design, sandboxed execution, analysis, documentation, and finalization) are described in \citet{liu2026autoresearchclaw}; here we focus on the discoveries it makes and the mechanisms that enable them.

\vspace{-0.5em}
\subsection{The Discovered Architecture}
\vspace{-0.5em}
\label{sec:method:arch}

\begin{figure}[t]
\centering
\includegraphics[width=\columnwidth]{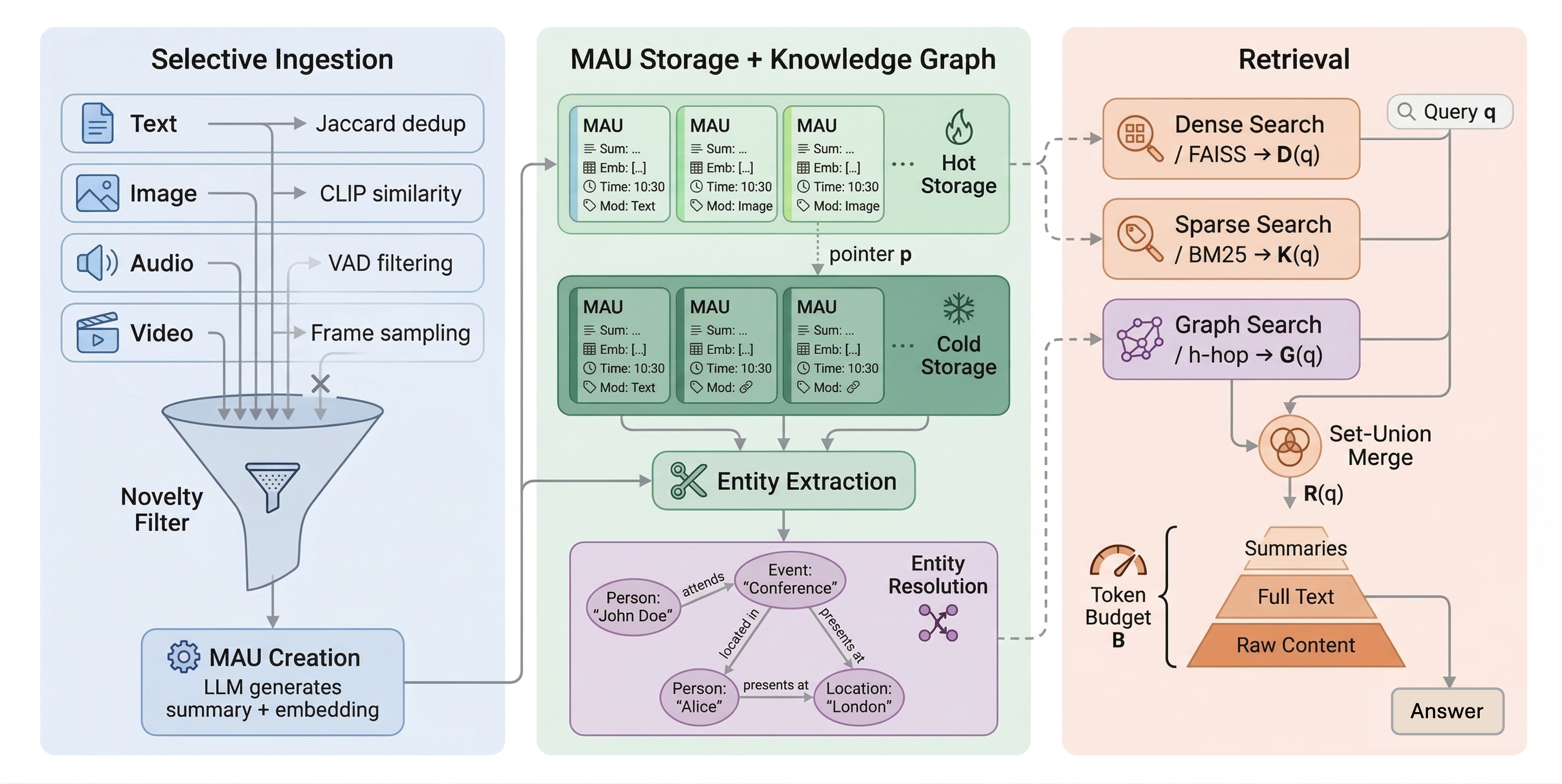}
\caption{\ours{} architecture overview. \textbf{Left:} Selective ingestion filters multimodal inputs (text, image, audio, video) via modality-specific novelty detectors and creates MAUs with LLM-generated summaries and embeddings. \textbf{Center:} MAUs are stored in hot storage (summaries, embeddings, metadata) and cold storage (raw content), with entity extraction building a knowledge graph with typed entities and relations. \textbf{Right:} Retrieval combines dense (FAISS), sparse (BM25), and graph ($h$-hop) search via set-union merging, then progressively expands results through a pyramid mechanism (summaries $\to$ full text $\to$ raw content) under a token budget $B$.}
\label{fig:architecture}
\vspace{-1.5em}
\end{figure}

The pipeline takes SimpleMem~\citep{liu2026simplemem}, a unimodal text-only lifelong memory framework, as its starting point.
We provide \tool{} with the SimpleMem codebase and instruct it to extend the system from text-only memory to full multimodal support, autonomously designing the necessary architectural components for ingesting, storing, and retrieving heterogeneous signals (text, images, audio, video).
Through iterative experimentation, the pipeline converges to an architecture organized around three principles: selective ingestion, progressive retrieval, and structured knowledge (Figure~\ref{fig:architecture}).

\vspace{-0.5em}
\subsubsection{Selective Ingestion}
\vspace{-0.5em}
\label{sec:method:ingestion}

The first principle is \textbf{selective ingestion}: the system first filters redundant inputs, then encapsulates the retained signals into a unified multimodal representation.

\noindent \textbf{Novelty-Based Filtering.}
Before any data enters the memory store, lightweight perceptual encoders assess the novelty of incoming information and discard redundant content.
For vision, CLIP embeddings are compared across consecutive frames to detect scene changes; for audio, VAD speech probability gates retention to reject silence; for text, Jaccard overlap with recent summaries filters near-duplicates.
This filtering significantly reduces storage requirements without losing semantic content.

\noindent \textbf{Multimodal Atomic Units.}
Signals that pass the novelty filter are encapsulated as \textbf{Multimodal Atomic Units (MAUs)}, $\mathcal{M} = \langle s, \mathbf{e}, p, \tau, m, \boldsymbol{\ell} \rangle$, which decouple compact searchable metadata from heavy raw content. Here $s$ is a text summary, $\mathbf{e}\in\mathbb{R}^d$ is its embedding, $p$ points to raw content in cold storage, $\tau$ is the timestamp, $m$ is modality, and $\boldsymbol{\ell}$ stores structural links to other MAUs.
This yields a two-tier design: \emph{hot storage} keeps summaries, embeddings, and temporal/graph metadata for fast retrieval, while \emph{cold storage} keeps large assets (images, audio, video) and is accessed lazily through $p$.

\subsubsection{Progressive Retrieval with Hybrid Search}
\label{sec:method:retrieval}

Once memories are ingested and stored as MAUs, the next challenge is how to retrieve them efficiently at query time. The second principle is \textbf{progressive retrieval}: rather than loading all retrieved content into the LLM context at once, \ours{} expands information in stages under explicit token budgets.

\noindent \textbf{Hybrid Dense-Sparse Search.}
Given a user query $q$, dense retrieval via FAISS~\citep{johnson2019billion}, a library for efficient similarity search over high-dimensional vectors, yields semantically similar candidates $\mathcal{D}(q)$ by inner-product search over L2-normalized MAU embeddings. In parallel, BM25~\citep{robertson2009bm25} scoring over MAU summaries yields keyword-matched candidates $\mathcal{K}(q)$.
A key discovery of the autonomous pipeline is set-union merging: empirically, score-based re-ranking (the standard approach) disrupts semantic ordering and degrades performance.
Instead, dense results retain their original ranking and BM25-only results are appended:
\begin{equation}
\label{eq:set_union}
    \mathcal{R}(q) = \mathcal{D}(q) \;\cup\; \big(\mathcal{K}(q) \setminus \mathcal{D}(q)\big).
\end{equation}

\noindent \textbf{Pyramid Retrieval.}
The hybrid search above produces a candidate set $\mathcal{R}(q)$, each candidate scored once by cosine similarity $\text{sim}(q, \mathcal{M}_i) = \mathbf{e}_q^\top \mathbf{e}_i$ during dense retrieval. The pyramid mechanism then progressively expands the \emph{content} of these candidates across three levels, reusing this score to gate each transition:
\textbf{Level~1} returns only summaries (${\sim}$10 tokens each) for the top-$k$ candidates by similarity;
\textbf{Level~2} loads full text or detailed captions for candidates whose similarity exceeds a threshold $\theta$;
\textbf{Level~3} loads raw content (images, audio) from cold storage under an explicit token budget B, expanding items greedily in decreasing similarity-per-token order.
All transitions are governed by deterministic rules rather than LLM judgment, avoiding additional latency while adapting context depth to each query's complexity.

\subsubsection{Knowledge Graph-Augmented Retrieval}
\label{sec:method:kg}

While hybrid search and pyramid retrieval handle queries that can be answered from individual MAUs, many real-world queries require reasoning over multiple connected facts (\eg, ``What gift did I give to the person I met at the conference in March?''). The third principle is therefore \textbf{structured knowledge}: \ours{} maintains a knowledge graph $\mathcal{G} = (\mathcal{V}, \mathcal{E})$ that captures entities and relationships across all MAUs.

During MAU creation, an LLM extracts entities and directed relations from each summary, producing entity-relation triples. Each entity carries a type label from 7 categories (Person, Location, Event, Concept, Time, Organization, Object) and is linked back to its source MAU. As new MAUs are ingested, the same real-world entity may appear under different surface forms (\eg, ``Dr.\ Smith'' vs.\ ``John Smith''). To prevent node fragmentation, entity resolution merges entities whose hybrid similarity, combining cosine similarity over name embeddings with Jaro-Winkler string similarity, exceeds a threshold.

At query time, the system identifies seed entities $\mathcal{V}_q \subset \mathcal{V}$ mentioned in the query and performs bounded neighborhood expansion within $h$ hops. Each reached entity is scored with distance-decayed relevance $r_{\mathcal{G}}(v) = \beta^{d(v, \mathcal{V}_q)} \cdot \text{conf}(v)$, where $d(v, \mathcal{V}_q)$ is the shortest path distance to any seed entity and $\beta \in (0,1)$ is a decay factor. MAUs linked to high-scoring graph entities are merged with the hybrid search results from $\mathcal{R}(q)$, providing both direct content matches and relationally connected evidence for answer generation.

\vspace{-0.5em}
\subsection{Benchmark-Specific Optimization}
\vspace{-0.5em}
\label{sec:method:optimization}

Having described the multimodal architecture that the pipeline discovers from SimpleMem, we now turn to how it optimizes this architecture for each target benchmark. The pipeline employs a two-phase strategy: rapid iteration on a small training subset, followed by evaluation on the held-out test set.

\noindent \textbf{Development subset for fast iteration.}
For each benchmark, the pipeline selects a small representative subset for rapid experimentation during the optimization loop. On LoCoMo, a small subset of conversations is used for iterative development, enabling each experiment to complete in under 2 hours. On Mem-Gallery, a small subset of datasets is used, with each experiment completing in minutes. This design enables the pipeline to explore dozens of hypotheses within days. After the optimization trajectory converges, the final configuration is evaluated on the complete benchmark to ensure generalization and to maintain consistency with the evaluation protocols used by prior memory systems.

\noindent \textbf{Iterative diagnosis and repair.}
During each optimization cycle, the pipeline autonomously diagnoses and repairs failures at two levels. At the \emph{execution level}, when an experiment fails or produces unexpected outputs, a self-healing module classifies the error (API error, dependency error, runtime exception, output format mismatch) and generates a targeted fix. For example, when the embedding service returned 403 errors due to an expired API key, the module detected the authentication failure pattern and switched to a local sentence-transformer backend without manual intervention.
At the \emph{semantic level}, when experiments succeed but produce unexpectedly poor metrics, the pipeline performs deeper analysis. 

% Two examples illustrate:
% \begin{enumerate}[nosep,leftmargin=*]
%     \item \textbf{Verbosity bug} (LoCoMo Iter~1): F1=0.322 despite correct retrieval. The pipeline found prediction length averaged 39.1 words versus 4.5 in gold answers, traced the issue to a missing \texttt{response\_format} parameter, and added the one-line fix (+175\%).
%     \item \textbf{Timestamp corruption} (LoCoMo Iter~5): 35\% of temporal answers contained the ingestion date rather than conversation dates. The pipeline generated a keyword-matching script that corrected 4,277 MAU timestamps with 99.98\% accuracy.
% \end{enumerate}

\vspace{-0.8em}
\section{Experiments}
\vspace{-0.8em}
\label{sec:experiments}

We evaluate \ours{} along two dimensions: (1)~the autonomous optimization process, specifically whether the pipeline discovers meaningful improvements across diverse benchmarks, and (2)~final system quality, examining whether the discovered architecture achieves state-of-the-art results and whether individual components contribute meaningfully.

\FloatBarrier  
\subsection{Experimental Setup}
\begin{figure*}[ht!]
  \centering
  \vspace{-0.5em}

  \begin{subfigure}{1\textwidth}
    \centering
    \includegraphics[width=\textwidth]{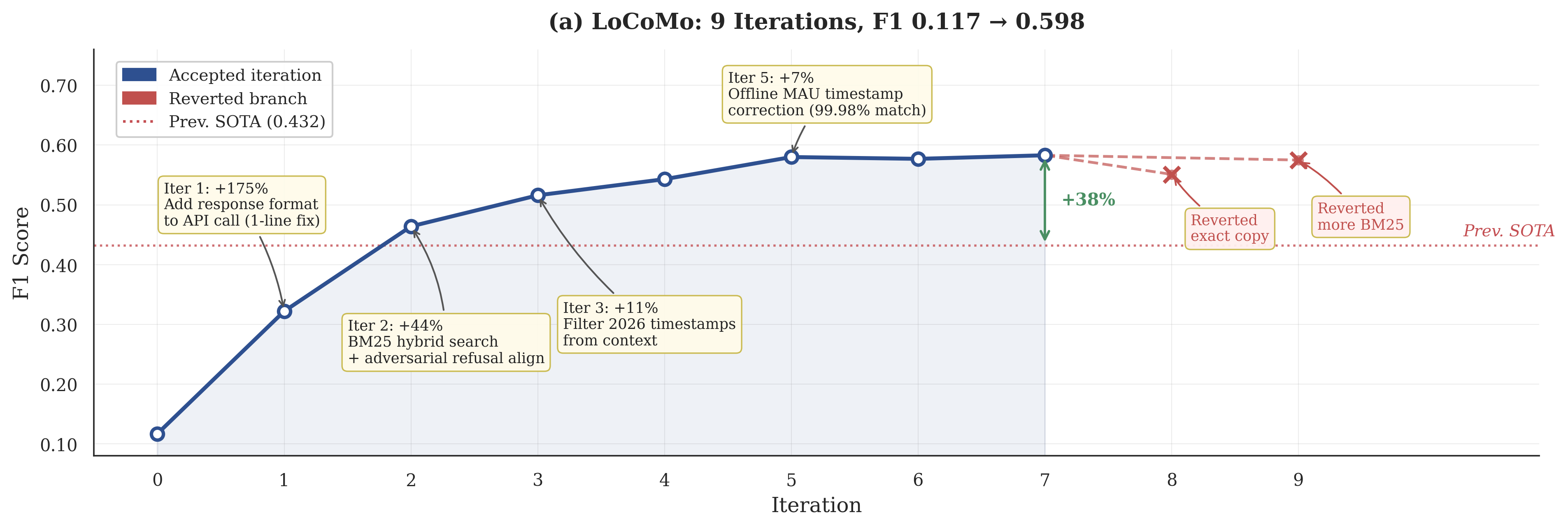}
  \end{subfigure}

  \begin{subfigure}{1\textwidth}
    \centering
    \includegraphics[width=\textwidth]{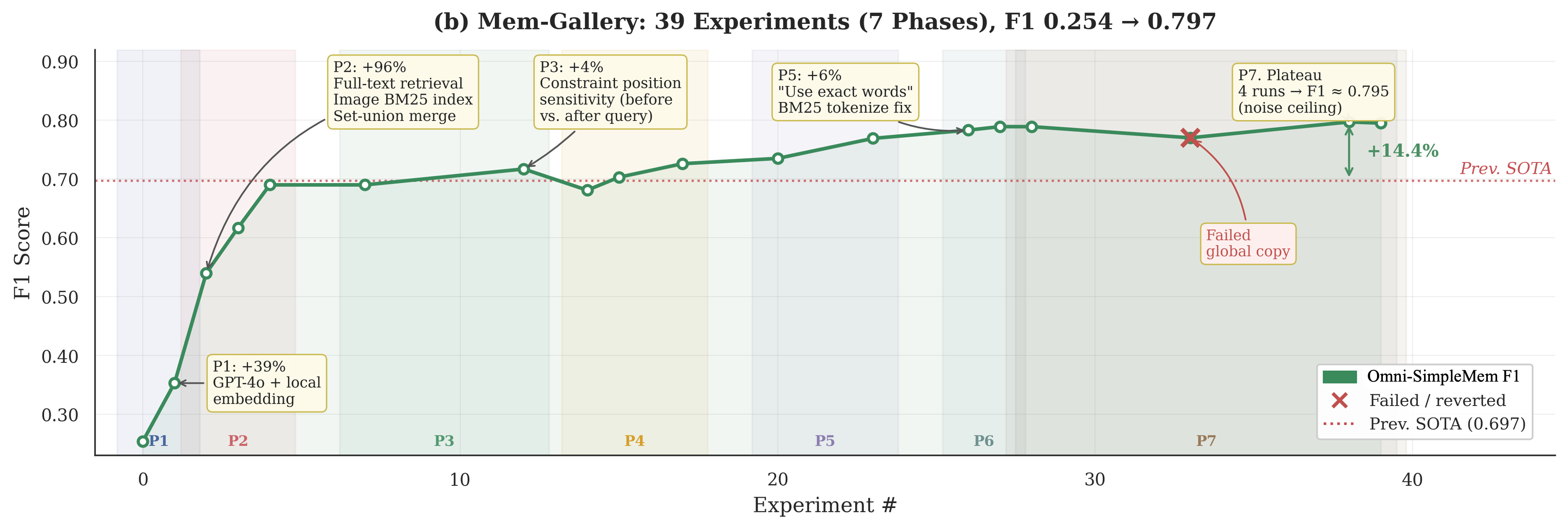}
  \end{subfigure}
    \vspace{-0.5em}
  \caption{Optimization trajectories on LoCoMo (top, 9 iterations)
    and Mem-Gallery (bottom, 39 experiments across 7 phases).
    Solid lines denote accepted iterations; failed/reverted experiments
    are marked with $\times$. The dashed red line indicates the previous
    SOTA\@. Key discoveries at each stage are annotated.}
  \label{fig:trajectories}
  \vspace{-2em}
\end{figure*}
% \begin{figure*}[t]
% \centering
% \includegraphics[width=\textwidth]{figures/fig1_optimization_trajectories.pdf}
% \caption{Optimization trajectories on LoCoMo (left, 9 iterations) and Mem-Gallery (right, 39 experiments across 7 phases). Successful iterations are shown as solid lines; failed/reverted experiments are marked with $\times$. The dashed red line indicates the previous SOTA. Key discoveries at each stage are annotated.}
% \label{fig:trajectories}
% \vspace{-1.5em}
% \end{figure*}

\label{sec:exp:setup}

\noindent \textbf{Benchmarks.}
We evaluate on two benchmarks spanning complementary types of memory-dependent reasoning: \textbf{LoCoMo}~\citep{locomo2024} (1,986 QA pairs across multi-session dialogues, token-level F1) and \textbf{Mem-Gallery}~\citep{memgallery2024} (1,711 QA pairs from 240 multimodal dialogues with 1,003 grounded images, F1). Detailed benchmark descriptions are provided in Appendix~\ref{app:benchmarks}.

\noindent \textbf{Baselines.}
We compare against six memory systems representing diverse design philosophies: \textbf{MemVerse}~\citep{memverse2024} (hierarchical episodic-semantic memory with multimodal knowledge graph), \textbf{Mem0}~\citep{mem02024} (dynamic fact extraction with optional graph memory), \textbf{Claude-Mem}~\citep{anthropic2024claude} (commercial embedding-based dialogue memory), \textbf{A-MEM}~\citep{xu2025amem} (LLM-directed memory reorganization), \textbf{MemGPT}~\citep{packer2023memgpt} (OS-inspired memory hierarchies), and \textbf{SimpleMem}~\citep{liu2026simplemem} (efficient lifelong memory with atomization and adaptive pruning). All are evaluated under identical splits and protocols.

\noindent \textbf{Implementation.}
Dense retrieval uses FAISS with all-MiniLM-L6-v2 embeddings (384d); sparse retrieval uses BM25; visual novelty filtering uses frozen CLIP ViT-B/32. Knowledge graph extraction uses GPT-4o in JSON mode. Default configuration: top-$k$=20, $\theta$=0.4, $B$=6,000 tokens. Full per-benchmark configuration details are provided in Appendix~\ref{app:implementation}.

\subsection{Optimization Trajectories}
\label{sec:exp:trajectories}

\begin{table*}[t]
\centering
\caption{Comparison across five LLM backbones on LoCoMo (left) and Mem-Gallery (right). LoCoMo columns: MH = multi-hop, SH = single-hop, Tmp = temporal, Open = open-domain, Adv = adversarial, All = overall F1. Mem-Gallery columns: F1, EM = exact match, B/B-1/B-2 = BLEU/BLEU-1/BLEU-2. Best baseline (excl.\ \ours{}) is \underline{underlined}; best overall is \textbf{bold}.}
\label{tab:locomo_comparison}
\label{tab:memgallery_comparison}
\small
\resizebox{\textwidth}{!}{%
\begin{tabular}{ll  cccccc  ccccc}
\toprule
& & \multicolumn{6}{c}{\textbf{LoCoMo}} & \multicolumn{5}{c}{\textbf{Mem-Gallery}} \\
\cmidrule(lr){3-8} \cmidrule(lr){9-13}
\textbf{Backbone} & \textbf{Method} & \textbf{MH} & \textbf{SH} & \textbf{Tmp} & \textbf{Open} & \textbf{Adv} & \textbf{All} & \textbf{F1} & \textbf{EM} & \textbf{B} & \textbf{B-1} & \textbf{B-2} \\
\midrule
\multirow{7}{*}{GPT-4o}
& MemVerse & 0.260 & 0.157 & 0.196 & 0.192 & \textbf{0.944} & 0.365 & 0.505 & 0.330 & 0.270 & 0.440 & 0.355 \\
& Mem0 & 0.309 & 0.156 & 0.217 & 0.295 & 0.857 & 0.397 & 0.298 & 0.192 & 0.182 & 0.268 & 0.224 \\
& Claude-Mem & 0.294 & 0.153 & 0.167 & 0.243 & \underline{0.915} & 0.383 & 0.210 & 0.148 & 0.148 & 0.194 & 0.170 \\
& A-MEM & 0.295 & 0.174 & 0.200 & 0.266 & 0.898 & 0.394 & 0.370 & 0.252 & 0.240 & 0.332 & 0.285 \\
& MemGPT & 0.305 & 0.188 & \underline{0.246} & 0.305 & 0.843 & 0.404 & 0.435 & 0.298 & 0.275 & 0.390 & 0.335 \\
& SimpleMem & \underline{0.318} & \underline{0.195} & 0.235 & \underline{0.308} & 0.802 & \underline{0.432} & \underline{0.535} & \underline{0.348} & \underline{0.310} & \underline{0.468} & \underline{0.390} \\
& \ours{} & \textbf{0.556} & \textbf{0.365} & \textbf{0.255} & \textbf{0.641} & 0.835 & \textbf{0.598} & \textbf{0.797} & \textbf{0.449} & \textbf{0.366} & \textbf{0.627} & \textbf{0.505} \\
\midrule
\multirow{7}{*}{GPT-4o-mini}
& MemVerse & 0.147 & 0.074 & 0.106 & 0.093 & 0.747 & 0.290 & 0.450 & 0.295 & 0.248 & 0.395 & 0.330 \\
& Mem0 & 0.285 & 0.112 & \underline{0.179} & 0.297 & 0.761 & 0.364 & 0.291 & 0.188 & 0.185 & 0.265 & 0.223 \\
& Claude-Mem & 0.245 & 0.102 & 0.122 & 0.215 & \underline{0.845} & 0.338 & 0.272 & 0.175 & 0.172 & 0.245 & 0.210 \\
& A-MEM & 0.278 & 0.091 & 0.163 & 0.260 & 0.823 & 0.357 & 0.330 & 0.222 & 0.205 & 0.298 & 0.252 \\
& MemGPT & 0.283 & 0.113 & \textbf{0.182} & 0.289 & 0.776 & 0.364 & 0.398 & 0.262 & 0.242 & 0.355 & 0.298 \\
& SimpleMem & \underline{0.300} & \underline{0.128} & 0.178 & \underline{0.312} & \textbf{0.891} & \underline{0.404} & \underline{0.498} & \underline{0.318} & \underline{0.290} & \underline{0.435} & \underline{0.368} \\
& \ours{} & \textbf{0.544} & \textbf{0.196} & 0.177 & \textbf{0.588} & 0.779 & \textbf{0.519} & \textbf{0.749} & \textbf{0.403} & \textbf{0.334} & \textbf{0.583} & \textbf{0.465} \\
\midrule
\multirow{7}{*}{GPT-4.1-nano}
& MemVerse & 0.146 & 0.061 & 0.169 & 0.115 & 0.711 & 0.256 & 0.470 & 0.308 & 0.255 & 0.410 & 0.340 \\
& Mem0 & 0.290 & 0.134 & 0.194 & 0.277 & 0.537 & 0.310 & 0.268 & 0.176 & 0.156 & 0.238 & 0.199 \\
& Claude-Mem & 0.087 & 0.029 & 0.119 & 0.047 & 0.705 & 0.246 & 0.303 & 0.194 & 0.172 & 0.268 & 0.223 \\
& A-MEM & 0.045 & 0.016 & 0.142 & 0.050 & \textbf{0.747} & 0.216 & 0.365 & 0.242 & 0.225 & 0.325 & 0.275 \\
& MemGPT & 0.287 & 0.130 & \underline{0.234} & 0.279 & 0.556 & 0.316 & 0.360 & 0.238 & 0.218 & 0.318 & 0.268 \\
& SimpleMem & \underline{0.298} & \underline{0.145} & 0.210 & \underline{0.285} & 0.648 & \underline{0.342} & \underline{0.518} & \underline{0.338} & \underline{0.300} & \underline{0.452} & \underline{0.380} \\
& \ours{} & \textbf{0.477} & \textbf{0.216} & \textbf{0.244} & \textbf{0.583} & \underline{0.722} & \textbf{0.492} & \textbf{0.780} & \textbf{0.430} & \textbf{0.353} & \textbf{0.610} & \textbf{0.488} \\
\midrule
\multirow{7}{*}{GPT-5.1}
& MemVerse & 0.287 & 0.173 & \underline{0.277} & 0.297 & 0.780 & 0.383 & 0.478 & 0.312 & 0.262 & 0.418 & 0.345 \\
& Mem0 & 0.292 & 0.160 & 0.261 & 0.298 & \underline{0.819} & 0.390 & 0.270 & 0.175 & 0.157 & 0.240 & 0.200 \\
& Claude-Mem & 0.289 & 0.171 & 0.264 & 0.292 & 0.814 & 0.388 & 0.305 & 0.203 & 0.188 & 0.279 & 0.230 \\
& A-MEM & 0.287 & 0.164 & 0.246 & 0.284 & \textbf{0.826} & 0.385 & 0.408 & 0.268 & 0.242 & 0.365 & 0.302 \\
& MemGPT & 0.288 & 0.165 & 0.249 & 0.294 & 0.806 & 0.385 & 0.425 & 0.275 & 0.250 & 0.378 & 0.315 \\
& SimpleMem & \underline{0.305} & \underline{0.178} & 0.272 & \underline{0.305} & 0.807 & \underline{0.418} & \underline{0.538} & \underline{0.350} & \underline{0.312} & \underline{0.470} & \underline{0.395} \\
& \ours{} & \textbf{0.598} & \textbf{0.367} & \textbf{0.307} & \textbf{0.676} & 0.747 & \textbf{0.613} & \textbf{0.810} & \textbf{0.460} & \textbf{0.374} & \textbf{0.639} & \textbf{0.515} \\
\midrule
\multirow{7}{*}{GPT-5-nano}
& MemVerse & 0.208 & \underline{0.203} & 0.168 & 0.252 & 0.741 & 0.366 & 0.478 & 0.315 & 0.262 & 0.420 & 0.345 \\
& Mem0 & 0.264 & 0.143 & 0.237 & 0.270 & 0.737 & 0.352 & 0.283 & 0.176 & 0.165 & 0.250 & 0.210 \\
& Claude-Mem & 0.091 & 0.063 & 0.092 & 0.088 & 0.736 & 0.275 & 0.350 & 0.249 & 0.217 & 0.315 & 0.264 \\
& A-MEM & 0.260 & 0.149 & 0.223 & 0.257 & \underline{0.745} & 0.348 & 0.505 & 0.332 & 0.290 & 0.445 & 0.368 \\
& MemGPT & 0.267 & 0.151 & 0.226 & 0.271 & 0.744 & 0.355 & 0.388 & 0.255 & 0.230 & 0.345 & 0.288 \\
& SimpleMem & \underline{0.278} & 0.200 & \underline{0.245} & \underline{0.282} & \textbf{0.824} & \underline{0.388} & \underline{0.522} & \underline{0.340} & \underline{0.302} & \underline{0.458} & \underline{0.385} \\
& \ours{} & \textbf{0.357} & \textbf{0.371} & \textbf{0.253} & \textbf{0.561} & 0.719 & \textbf{0.522} & \textbf{0.787} & \textbf{0.437} & \textbf{0.357} & \textbf{0.617} & \textbf{0.494} \\
\bottomrule
\end{tabular}%
}
\vspace{-2em}
\end{table*}

Figure~\ref{fig:trajectories} visualizes the optimization trajectories. The pipeline completed ${\sim}50$ experiments across both benchmarks in ${\sim}$72 hours of wall-clock time, a coverage that would require approximately 4 weeks for a human researcher at ${\sim}$3 experiments per day. We highlight the most impactful discoveries; detailed per-iteration tables are provided in Appendix~\ref{app:iteration_logs}.

\noindent \textbf{LoCoMo (9 iterations, F1: 0.117$\to$0.598).}
The pipeline executed 9 successful iterations over 48 hours, with 2 additional experiments automatically reverted (full trajectory in Appendix Table~\ref{tab:locomo_trajectory}).
The most impactful discovery (Iter~1, +175\%) was identifying that the API call lacked a \texttt{response\_format} parameter, a one-line bug causing 9$\times$ verbosity that destroyed F1 precision.
In Iter 5, the pipeline discovered that all 4,277 MAU timestamps had been corrupted to the ingestion date and autonomously generated a keyword-matching script that corrected 99.98\% of them without re-ingestion.
The pipeline also discovered that set-union merging of FAISS and BM25 results (Iter~2) significantly outperforms score-based fusion, a finding confirmed by ablation (Section~\ref{sec:exp:ablation}).

\noindent \textbf{Mem-Gallery (39 experiments, F1: 0.254$\to$0.797).}
Optimization spanned 7 phases (full trajectory in Appendix Table~\ref{tab:memgallery_trajectory}).
The single largest improvement (+53\%) came from discovering that returning full original dialogue text instead of LLM-generated summaries dramatically improves token-overlap F1, a non-obvious finding, since summaries are traditionally preferred for efficiency.
The pipeline also found that prompt constraint \emph{positioning} (before vs.\ after the question) matters more than constraint \emph{content}, with one category improving +188\% from this change alone.
After Phase~7, four independent runs yielded F1 in [0.791, 0.797], confirming the performance ceiling and triggering the pipeline's decision to stop.

\subsection{Main Results}
\label{sec:exp:results}

To contextualize these results against existing memory systems, we conduct a controlled comparison of \ours{} against six baselines across five LLM backbones (GPT-4o, GPT-4o-mini, GPT-4.1-nano, GPT-5.1, and GPT-5-nano).
Table~\ref{tab:locomo_comparison} reports per-category F1 on LoCoMo and five evaluation metrics on Mem-Gallery across all backbones.

On LoCoMo, \ours{} achieves the highest overall F1 across all backbones, ranging from 0.492 (GPT-4.1-nano) to 0.613 (GPT-5.1), substantially outperforming SimpleMem (0.342--0.432), the current state-of-the-art on LoCoMo~\citep{liu2026simplemem}.
\ours{} dominates on multi-hop, single-hop, and open-domain categories, with particularly large margins on open-domain questions.

On Mem-Gallery, \ours{} achieves F1 ranging from 0.749 to 0.810, consistently outperforming all memory baselines by a wide margin.
SimpleMem is again the strongest baseline (F1 up to 0.538 with GPT-5.1), but still trails \ours{} by over 25 percentage points.
These patterns confirm that \ours{}'s gains come from its architectural design (hybrid search, pyramid retrieval, knowledge graph augmentation) rather than from a single dominant component.

\subsection{Analysis}
\label{sec:exp:analysis}

\subsubsection{Ablation Studies}
\label{sec:exp:ablation}

\begin{wraptable}{r}{0.45\columnwidth}
\vspace{-2.5em}
\centering
\caption{Component ablation on LoCoMo (mean $\Delta$F1$\times$100 across 4 backbones).}
\label{tab:ablations}
\small
\begin{tabular}{@{}lcc@{}}
\toprule
\textbf{Component Removed} & \textbf{$\Delta$F1} & \textbf{Rel.} \\
\midrule
w/o Pyramid Expansion & $-$10.2 & $-$17\% \\
w/o BM25 Hybrid & $-$8.5 & $-$14\% \\
w/o LLM Summarization & $-$7.3 & $-$12\% \\
Reduced top-$k$ (5 vs 20) & $-$4.2 & $-$7\% \\
w/o Metadata Context & $-$1.4 & $-$2\% \\
\bottomrule
\end{tabular}
\vspace{-1em}
\end{wraptable}
Table 2 presents an ablation study on LoCoMo that validates key design choices discovered by the pipeline. Specifically, we remove individual components and report the mean $\Delta$F1 across 4 LLM backbones. Pyramid expansion ($-$17\%) and BM25 hybrid search ($-$14\%) are the most critical, and both were significantly refined by the autonomous pipeline during optimization. LLM summarization contributes $-$12\%, confirming that compact MAU summaries are essential for retrieval quality. Reducing top-$k$ from 20 to 5 costs $-$7\%, while metadata context has a modest effect ($-$2\%). Notably, the two most impactful components (pyramid expansion and hybrid search) are precisely those that received the most optimization iterations, suggesting the pipeline correctly allocated its search budget.

% \noindent \textbf{Retrieval Strategy (Table~\ref{tab:ablations}, left).}
% The set-union merging approach discovered by the pipeline outperforms all alternatives. Using dense retrieval alone achieves F1\,=\,0.516; adding sparse retrieval via score fusion improves overall F1 to 0.540 but degrades multi-hop reasoning (0.441$\to$0.390), because re-ranking by combined scores disrupts the semantic ordering established by dense retrieval. Set-union merging avoids this trade-off by preserving the dense ranking while appending BM25-only results, achieving the best aggregate F1 (0.583) without sacrificing any category.

% \noindent \textbf{Component Ablation (Table~\ref{tab:ablations}, right).}

\begin{wrapfigure}{r}{0.45\textwidth}
\vspace{-5em}
\centering
\includegraphics[width=0.46\textwidth]{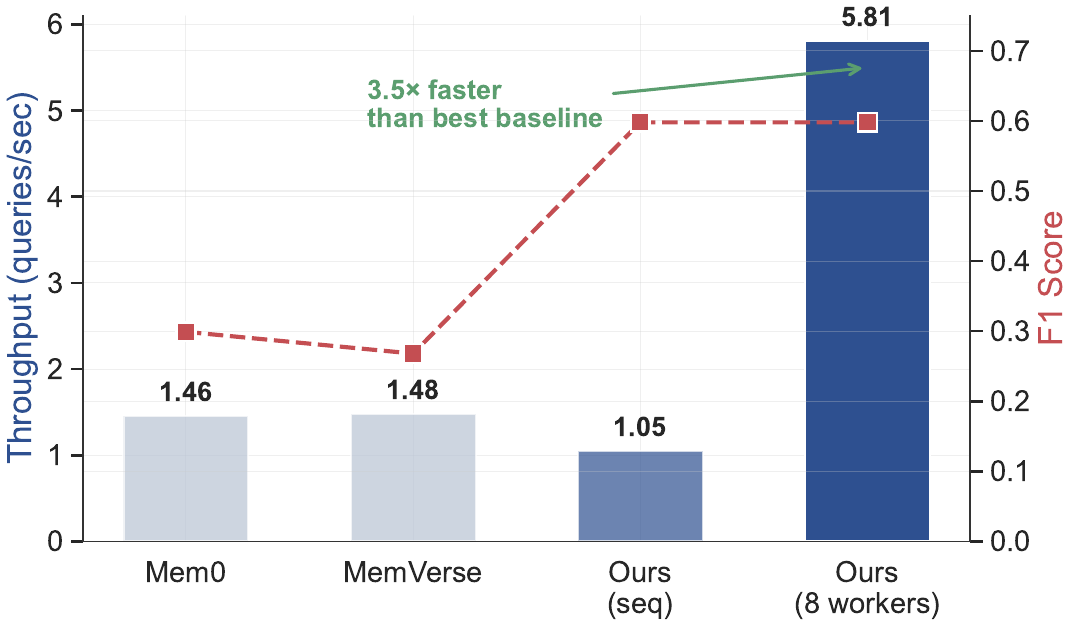}
\caption{Throughput vs.\ F1. \ours{} with 8 workers achieves 3.5$\times$ higher throughput.}
\label{fig:efficiency}
\vspace{-1em}
\end{wrapfigure}

\subsubsection{Efficiency}
\label{sec:exp:efficiency}

\ours{} achieves 5.81 queries/sec with 8 parallel workers (3.5$\times$ faster than the fastest baseline), enabled by read-only FAISS and BM25 indices supporting concurrent lookup (Figure~\ref{fig:efficiency}, Table~\ref{tab:efficiency}).
All baselines are bottlenecked by sequential LLM generation (85--97\% of per-query time), while \ours{} parallelizes the retrieval-generation pipeline via thread-safe read-only indices.

\begin{wraptable}{r}{0.5\columnwidth}
\vspace{-1.5em}
\centering
\caption{Latency breakdown.}
\label{tab:efficiency}
\small
\begin{tabular}{@{}lccc@{}}
\toprule
\textbf{Method} & \textbf{q/s} & \textbf{Ret.} & \textbf{Gen.} \\
\midrule
SimpleMem & 1.68 & 45 & 550 \\
MemVerse & 1.48 & 70 & 596 \\
Mem0 & 1.46 & 18 & 665 \\
\ours{} (w=1) & 1.05 & 118 & 846 \\
\ours{} (w=8) & \textbf{5.81} & 461 & 821 \\
\bottomrule
\multicolumn{4}{@{}l}{\scriptsize Ret./Gen.\ in milliseconds.}
\end{tabular}
\vspace{-1em}
\end{wraptable}

\subsubsection{Case Study: Multi-Hop Retrieval}
\label{sec:exp:case_study}

We illustrate \ours{}'s retrieval pipeline on a real multi-hop query from LoCoMo that requires synthesizing facts across separate conversation sessions. The query asks: \emph{``What subject have Caroline and Melanie both painted?''} The gold answer is ``sunsets,'' but answering correctly requires retrieving each person's painting history from different sessions and identifying the overlap.

\noindent \textbf{Hybrid search.}
Dense retrieval returns MAUs mentioning Caroline's paintings (\eg, ``Caroline painted a sunset'') and Melanie's art projects (\eg, ``Mel and her kids painted a sunset with a tree''), but these appear in separate sessions with different surrounding context. BM25 recovers additional MAUs containing the keyword ``paint'' that rank lower in dense results. Set-union merging preserves the dense ordering and appends BM25-only matches.

\noindent \textbf{Knowledge graph expansion.}
The query processor extracts seed entities \textsc{Caroline} (Person) and \textsc{Melanie} (Person). Neighborhood expansion links both entities to \textsc{painting} (Concept) and \textsc{sunset} (Concept) through separate relation paths, surfacing MAUs that mention each person's painting activities even when the surface text does not co-mention both names.

\noindent \textbf{Pyramid retrieval and answer.}
Level~1 summaries from both relation paths are loaded; their similarity scores exceed $\theta$, triggering Level~2 expansion of the full conversation text. The LLM identifies ``sunsets'' as the common subject and produces the correct answer (F1\,=\,1.0). In contrast, MemGPT, which lacks cross-session entity linking, hallucinates ``Horses'' (F1\,=\,0.0).

\vspace{-0.5em}
\section{Conclusion}
\vspace{-0.5em}
\label{sec:conclusion}

We have presented \ours{}, a unified multimodal memory framework whose architecture and configuration are discovered through \tool{}, an autonomous research pipeline. Starting from a na\"ive baseline, the pipeline autonomously executed ${\sim}50$ experiments in ${\sim}$72 hours, achieving state-of-the-art on both LoCoMo and Mem-Gallery. The highest-impact discoveries, including bug fixes, architectural changes, and prompt engineering, require code comprehension and cross-component reasoning beyond the reach of traditional AutoML. Our taxonomy of six discovery types, together with the observation that multimodal memory is well-suited for autoresearch due to its scalar metrics, modular architecture, and fast iteration cycles, provides a roadmap for applying autonomous research pipelines to other complex AI system domains.

\bibliography{references}

@misc{lu2024aiscientist,
  title={The AI Scientist: Towards Fully Automated Open-Ended Scientific Discovery}, 
      author={Chris Lu and Cong Lu and Robert Tjarko Lange and Jakob Foerster and Jeff Clune and David Ha},
      year={2024},
      eprint={2408.06292},
      archivePrefix={arXiv},
      primaryClass={cs.AI},
      url={https://arxiv.org/abs/2408.06292}, 
}

@article{lu2025aiscientistv2,
  title={The AI Scientist-v2: Workshop-Level Automated Scientific Discovery via Agentic Tree Search}, 
      author={Yutaro Yamada and Robert Tjarko Lange and Cong Lu and Shengran Hu and Chris Lu and Jakob Foerster and Jeff Clune and David Ha},
      year={2025},
      eprint={2504.08066},
      archivePrefix={arXiv},
      primaryClass={cs.AI},
      url={https://arxiv.org/abs/2504.08066}, 
}

@article{romeraparedes2024funsearch,
  author = {Romera-Paredes, Bernardino and Barekatain, Mohammadamin and Novikov, Alexander and Balog, Matej and Kumar, M. Pawan and Dupont, Emilien and Ruiz, Francisco J. R. and Ellenberg, Jordan S. and Wang, Pengming and Fawzi, Omar and Kohli, Pushmeet and Fawzi, Alhussein},
  title = {Mathematical discoveries from program search with large language models},
  journal = {Nature},
  year = {2024},
  volume = {625},
  number = {7995},
  pages = {468--475},
  doi = {10.1038/s41586-023-06924-6},
  url = {https://doi.org/10.1038/s41586-023-06924-6}
}

@article{panfilov2026claudini,
  title={Claudini: Autoresearch Discovers State-of-the-Art Adversarial Attack Algorithms for LLMs}, 
      author={Alexander Panfilov and Peter Romov and Igor Shilov and Yves-Alexandre de Montjoye and Jonas Geiping and Maksym Andriushchenko},
      year={2026},
      eprint={2603.24511},
      archivePrefix={arXiv},
      primaryClass={cs.LG},
      url={https://arxiv.org/abs/2603.24511}, 
}

@article{tang2025airesearcher,
  title={AI-Researcher: Autonomous Scientific Innovation}, 
      author={Jiabin Tang and Lianghao Xia and Zhonghang Li and Chao Huang},
      year={2025},
      eprint={2505.18705},
      archivePrefix={arXiv},
      primaryClass={cs.AI},
      url={https://arxiv.org/abs/2505.18705}, 
}

@article{qu2026bilevel,
  title={Bilevel Autoresearch: Meta-Autoresearching Itself}, 
      author={Yaonan Qu and Meng Lu},
      year={2026},
      eprint={2603.23420},
      archivePrefix={arXiv},
      primaryClass={cs.AI},
      url={https://arxiv.org/abs/2603.23420},
}

@article{tie2025survey,
  title={A Survey of AI Scientists}, 
      author={Guiyao Tie and Pan Zhou and Lichao Sun},
      year={2026},
      eprint={2510.23045},
      archivePrefix={arXiv},
      primaryClass={cs.AI},
      url={https://arxiv.org/abs/2510.23045}, 
}

@misc{liu2026autoresearchclaw,
  author       = {Liu, Jiaqi and Xia, Peng and Han, Siwei and Qiu, Shi and Zhang, Letian and Chen, Guiming and Tu, Haoqin and Yang, Xinyu and Zhou, Jiawei and Zhu, Hongtu and Li, Yun and Zhou, Yuyin and Zheng, Zeyu and Xie, Cihang and Ding, Mingyu and Yao, Huaxiu},
  title        = {{AutoResearchClaw}: Fully Autonomous Research from Idea to Paper},
  year         = {2026},
  organization = {GitHub},
  url          = {https://github.com/aiming-lab/AutoResearchClaw},
}

@article{liu2026simplemem,
  title={SimpleMem: Efficient Lifelong Memory for LLM Agents}, 
      author={Jiaqi Liu and Yaofeng Su and Peng Xia and Siwei Han and Zeyu Zheng and Cihang Xie and Mingyu Ding and Huaxiu Yao},
      year={2026},
      eprint={2601.02553},
      archivePrefix={arXiv},
      primaryClass={cs.AI},
      url={https://arxiv.org/abs/2601.02553}, 
}

@inproceedings{packer2023memgpt,
  title={{MemGPT}: Towards LLMs as Operating Systems},
  author={Packer, Charles and Wooders, Sarah and Lin, Kevin and Fang, Vivian and Patil, Shishir G. and Stoica, Ion and Gonzalez, Joseph E.},
  journal={arXiv preprint arXiv:2310.08560},
  year={2023}
}

@inproceedings{park2023generative,
  title={Generative Agents: Interactive Simulacra of Human Behavior}, 
      author={Joon Sung Park and Joseph C. O'Brien and Carrie J. Cai and Meredith Ringel Morris and Percy Liang and Michael S. Bernstein},
      year={2023},
      eprint={2304.03442},
      archivePrefix={arXiv},
      primaryClass={cs.HC},
      url={https://arxiv.org/abs/2304.03442}, 
}

@article{xu2025amem,
  title={A-MEM: Agentic Memory for LLM Agents}, 
      author={Wujiang Xu and Zujie Liang and Kai Mei and Hang Gao and Juntao Tan and Yongfeng Zhang},
      year={2025},
      eprint={2502.12110},
      archivePrefix={arXiv},
      primaryClass={cs.CL},
      url={https://arxiv.org/abs/2502.12110}, 
}

@article{memverse2024,
  title={MemVerse: Multimodal Memory for Lifelong Learning Agents}, 
      author={Junming Liu and Yifei Sun and Weihua Cheng and Haodong Lei and Yirong Chen and Licheng Wen and Xuemeng Yang and Daocheng Fu and Pinlong Cai and Nianchen Deng and Yi Yu and Shuyue Hu and Botian Shi and Ding Wang},
      year={2025},
      eprint={2512.03627},
      archivePrefix={arXiv},
      primaryClass={cs.AI},
      url={https://arxiv.org/abs/2512.03627},
}

@article{mem02024,
  title={Mem0: Building Production-Ready AI Agents with Scalable Long-Term Memory}, 
      author={Prateek Chhikara and Dev Khant and Saket Aryan and Taranjeet Singh and Deshraj Yadav},
      year={2025},
      eprint={2504.19413},
      archivePrefix={arXiv},
      primaryClass={cs.CL},
      url={https://arxiv.org/abs/2504.19413},
}

@article{visrag2024,
  title={VisRAG: Vision-based Retrieval-augmented Generation on Multi-modality Documents}, 
      author={Shi Yu and Chaoyue Tang and Bokai Xu and Junbo Cui and Junhao Ran and Yukun Yan and Zhenghao Liu and Shuo Wang and Xu Han and Zhiyuan Liu and Maosong Sun},
      year={2025},
      eprint={2410.10594},
      archivePrefix={arXiv},
      primaryClass={cs.IR},
      url={https://arxiv.org/abs/2410.10594}, 
}

@misc{anthropic2024claude,
  title={Claude 3 Model Card},
  author={Anthropic},
  year={2024},
  howpublished={\url{https://www.anthropic.com}}
}

@article{locomo2024,
  title={Evaluating very long-term conversational memory of llm agents},
  author={Maharana, Adyasha and Lee, Dong-Ho and Tulyakov, Sergey and Bansal, Mohit and Barbieri, Francesco and Fang, Yuwei},
  journal={arXiv preprint arXiv:2402.17753},
  year={2024}
}

@article{memgallery2024,
  title={Mem-Gallery: Benchmarking Multimodal Long-Term Conversational Memory for MLLM Agents}, 
      author={Yuanchen Bei and Tianxin Wei and Xuying Ning and Yanjun Zhao and Zhining Liu and Xiao Lin and Yada Zhu and Hendrik Hamann and Jingrui He and Hanghang Tong},
      year={2026},
      eprint={2601.03515},
      archivePrefix={arXiv},
      primaryClass={cs.CL},
      url={https://arxiv.org/abs/2601.03515}, 
}

@book{hutter2019automl,
  title={Automated Machine Learning: Methods, Systems, Challenges},
  author={Hutter, Frank and Kotthoff, Lars and Vanschoren, Joaquin},
  year={2019},
  publisher={Springer}
}

@inproceedings{zoph2017nas,
  title={Neural Architecture Search with Reinforcement Learning},
  author={Zoph, Barret and Le, Quoc V},
  booktitle={International Conference on Learning Representations (ICLR)},
  year={2017}
}

@inproceedings{liu2019darts,
  title={{DARTS}: Differentiable Architecture Search},
  author={Liu, Hanxiao and Simonyan, Karen and Yang, Yiming},
  booktitle={International Conference on Learning Representations (ICLR)},
  year={2019}
}

@inproceedings{bergstra2011tpe,
  title={Algorithms for Hyper-Parameter Optimization},
  author={Bergstra, James and Bardenet, R{\'e}mi and Bengio, Yoshua and K{\'e}gl, Bal{\'a}zs},
  booktitle={Advances in Neural Information Processing Systems},
  year={2011}
}

@article{falkner2018bohb,
  title={{BOHB}: Robust and Efficient Hyperparameter Optimization at Scale},
  author={Falkner, Stefan and Klein, Aaron and Hutter, Frank},
  journal={International Conference on Machine Learning},
  year={2018}
}

@article{lewis2020rag,
  title={Retrieval-Augmented Generation for Knowledge-Intensive {NLP} Tasks},
  author={Lewis, Patrick and Perez, Ethan and Piktus, Aleksandra and Petroni, Fabio and Karpukhin, Vladimir and Goyal, Naman and K{\"u}ttler, Heinrich and Lewis, Mike and Yih, Wen-tau and Rockt{\"a}schel, Tim and others},
  journal={Advances in Neural Information Processing Systems},
  volume={33},
  pages={9459--9474},
  year={2020}
}

@article{robertson2009bm25,
  title={The Probabilistic Relevance Framework: {BM25} and Beyond},
  author={Robertson, Stephen and Zaragoza, Hugo},
  journal={Foundations and Trends in Information Retrieval},
  volume={3},
  number={4},
  pages={333--389},
  year={2009}
}

@article{johnson2019billion,
  title={Billion-Scale Similarity Search with {GPUs}},
  author={Johnson, Jeff and Douze, Matthijs and J{\'e}gou, Herv{\'e}},
  journal={IEEE Transactions on Big Data},
  volume={7},
  number={3},
  pages={535--547},
  year={2021}
}

@article{xu2025amemagenticmemoryllm,
  title={A-MEM: Agentic Memory for LLM Agents},
  author={Wujiang Xu and Zujie Liang and Kai Mei and Hang Gao and Juntao Tan and Yongfeng Zhang},
  year={2025},
  eprint={2502.12110},
  archivePrefix={arXiv},
  primaryClass={cs.CL},
  url={https://arxiv.org/abs/2502.12110},
}

@book{hutter2019automated,
  title={Automated Machine Learning: Methods, Systems, Challenges},
  author={Hutter, Frank and Kotthoff, Lars and Vanschoren, Joaquin},
  year={2019},
  publisher={Springer}
}

@inproceedings{yao2022react,
  title={{ReAct}: Synergizing Reasoning and Acting in Language Models},
  author={Yao, Shunyu and Zhao, Jeffrey and Yu, Dian and Du, Nan and Shafran, Izhak and Narasimhan, Karthik and Cao, Yuan},
  booktitle={International Conference on Learning Representations (ICLR)},
  year={2023}
}

@article{yan2025positionmultimodallargelanguage,
  title={Position: Multimodal Large Language Models Can Significantly Advance Scientific Reasoning}, 
      author={Yibo Yan and Shen Wang and Jiahao Huo and Jingheng Ye and Zhendong Chu and Xuming Hu and Philip S. Yu and Carla Gomes and Bart Selman and Qingsong Wen},
      year={2025},
      eprint={2502.02871},
      archivePrefix={arXiv},
      primaryClass={cs.CL},
      url={https://arxiv.org/abs/2502.02871}, 
}

@article{zhang2024surveymemorymechanismlarge,
  title={A Survey on the Memory Mechanism of Large Language Model based Agents}, 
      author={Zeyu Zhang and Xiaohe Bo and Chen Ma and Rui Li and Xu Chen and Quanyu Dai and Jieming Zhu and Zhenhua Dong and Ji-Rong Wen},
      year={2024},
      eprint={2404.13501},
      archivePrefix={arXiv},
      primaryClass={cs.AI},
      url={https://arxiv.org/abs/2404.13501}, 
}

@inproceedings{borgeaud2022retro,
  title={Improving Language Models by Retrieving from Trillions of Tokens},
  author={Borgeaud, Sebastian and Mensch, Arthur and Hoffmann, Jordan and Cai, Trevor and Rutherford, Eliza and Millican, Katie and van den Driessche, George and Lespiau, Jean-Baptiste and Damoc, Bogdan and Clark, Aidan and de Las Casas, Diego and Guy, Aurelia and Menick, Jacob and Ring, Roman and Hennigan, Tom and Huang, Saffron and Maggiore, Loren and Jones, Chris and Cassirer, Albin and Brock, Andy and Paganini, Michela and Irving, Geoffrey and Vinyals, Oriol and Osindero, Simon and Simonyan, Karen and Rae, Jack W. and Elsen, Erich and Sifre, Laurent},
  booktitle={Proceedings of the 39th International Conference on Machine Learning (ICML)},
  volume={162},
  pages={2206--2240},
  year={2022},
  publisher={PMLR}
}

@misc{feurer2022autosklearn20handsfreeautoml,
      title={Auto-Sklearn 2.0: Hands-free AutoML via Meta-Learning}, 
      author={Matthias Feurer and Katharina Eggensperger and Stefan Falkner and Marius Lindauer and Frank Hutter},
      year={2022},
      eprint={2007.04074},
      archivePrefix={arXiv},
      primaryClass={cs.LG},
      url={https://arxiv.org/abs/2007.04074}, 
}

@misc{huang2024mlagentbenchevaluatinglanguageagents,
      title={MLAgentBench: Evaluating Language Agents on Machine Learning Experimentation}, 
      author={Qian Huang and Jian Vora and Percy Liang and Jure Leskovec},
      year={2024},
      eprint={2310.03302},
      archivePrefix={arXiv},
      primaryClass={cs.LG},
      url={https://arxiv.org/abs/2310.03302}, 
}
\bibliographystyle{colm2026_conference}

\newpage
\appendix
%% ============================================================
%% A. Benchmark Details (corresponds to 4.1 Setup)
%% ============================================================
\section{Benchmark Details}
\label{app:benchmarks}

\paragraph{LoCoMo.}
The Long-term Conversation Memory benchmark \citep{locomo2024} evaluates an agent's ability to recall and reason over extended multi-session dialogues.
It comprises 10 conversations with 19--32 sessions each (${\sim}9$K tokens per dialogue), generating 1,986 QA pairs across five categories:
\textbf{Single-hop (SH)} requires retrieving a single fact;
\textbf{Multi-hop (MH)} requires synthesizing information across multiple sessions;
\textbf{Temporal (T)} tests reasoning about when events occurred;
\textbf{Open-ended (O)} requires generating longer, contextual responses; and
\textbf{Adversarial (A)} tests the ability to correctly refuse unanswerable questions.
Evaluation uses token-level F1 with Porter stemming.

\paragraph{Mem-Gallery.}
Mem-Gallery \citep{memgallery2024} evaluates multimodal long-term memory in social interactions, comprising 1,711 QA pairs from 240 multi-session dialogues with 1,003 grounded images and 3,962 conversational rounds.
Questions span 9 categories: Action Recognition (AR), Compound Decomposition (CD), Visual Search (VS), Timeline Learning (TTL), Temporal Reasoning (TR), Fact Retrieval (FR), Visual Reasoning (VR), Knowledge Reasoning (KR), and Multi-Entity Reasoning (MR).
We use the public 20-dataset subset from HuggingFace.

%% ============================================================
%% B. Baseline Descriptions (corresponds to 4.1 Setup)
%% ============================================================
\section{Baseline Descriptions}
\label{app:baselines}

\paragraph{MemVerse} \citep{memverse2024} combines episodic and semantic memory layers with a multimodal knowledge graph.
Each ingested item requires 3 LLM calls (summarization, entity extraction, relation linking), yielding an ingestion rate of 0.22 items/sec.
Retrieval uses embedding-based similarity over the knowledge graph.

\paragraph{Mem0} \citep{mem02024} performs dynamic fact extraction from conversational inputs, storing structured facts with optional graph memory augmentation.
It achieves 1.28 items/sec ingestion and 18ms search latency, but is fundamentally text-only.

\paragraph{SimpleMem} \citep{liu2026simplemem} introduces efficient lifelong memory management through three core mechanisms: memory atomization (decomposing inputs into fine-grained atomic units), adaptive consolidation (merging related memories to reduce redundancy), and context-aware pruning.
It achieves state-of-the-art performance on LoCoMo with significantly reduced token cost (${\sim}$45\% less than Mem0) and 4$\times$ faster total processing time.

\paragraph{Claude-Mem} \citep{anthropic2024claude} provides commercial embedding-based dialogue memory with per-turn image storage.

\paragraph{A-MEM} \citep{xu2025amem} introduces agentic memory management where the LLM itself decides when and how to reorganize stored memories, including operations such as merging, splitting, and abstracting memory entries to maintain a compact and relevant memory state.

\paragraph{MemGPT} \citep{packer2023memgpt} draws an analogy to operating system memory hierarchies, implementing a main context (analogous to RAM) and external storage (analogous to disk) with explicit \texttt{push}/\texttt{pop} operations managed by the LLM agent.
It achieves strong performance on conversational tasks but is fundamentally text-only and bottlenecked by sequential LLM calls for each memory management decision.

%% ============================================================
%% C. Implementation Details (corresponds to 4.1 Setup + Section 3)
%% ============================================================
\section{\ours{} Technical Details}
\label{app:omnimem_details}
\label{app:implementation}

\subsection{System Architecture}

The complete \ours{} implementation comprises 13,300 lines of Python across 11 subpackages:

\begin{itemize}[nosep,leftmargin=*]
    \item \textbf{Core} (MAU data structure, event hierarchy, unified configuration)
    \item \textbf{Processors} (modality-specific ingestion: text, image, audio, video)
    \item \textbf{Storage} (three-layer persistence: MAU store as JSON Lines, FAISS vector store, cold storage on filesystem/S3)
    \item \textbf{Retrieval} (pyramid retriever, query processor, expansion manager)
    \item \textbf{Knowledge} (entity extractor, in-memory knowledge graph, graph retriever)
    \item \textbf{Orchestrator} (central coordinator, 873 lines, unified API)
\end{itemize}

\subsection{Hyperparameters and Benchmark-Specific Configurations}

Table~\ref{tab:hyperparams} lists the complete hyperparameter specification. Table~\ref{tab:benchmark_configs} summarizes the per-benchmark configurations discovered by the pipeline.

\begin{table}[h]
\centering
\small
\caption{Hyperparameter specification of \ours{}.}
\label{tab:hyperparams}
\begin{tabular}{@{}llc@{}}
\toprule
\textbf{Component} & \textbf{Parameter} & \textbf{Value} \\
\midrule
\multirow{3}{*}{Visual Ingestion} & CLIP threshold $\tau_{\mathrm{high}}$ & 0.9 \\
 & CLIP threshold $\tau_{\mathrm{low}}$ & 0.7 \\
 & Frame buffer size & 3 \\
\midrule
Audio Ingestion & VAD threshold & 0.5 \\
Text Ingestion & Jaccard threshold $\tau_{\mathrm{dup}}$ & 0.8 \\
\midrule
\multirow{4}{*}{Entity Resolution} & Merge threshold $\tau_{\mathrm{res}}$ & 0.85 \\
 & Semantic weight $\alpha$ & 0.5 \\
 & Graph decay $\beta$ & 0.7 \\
 & Expansion hops $h$ & 2 \\
\midrule
\multirow{3}{*}{Pyramid Retrieval} & Auto-expand threshold $\theta$ & 0.4 \\
 & Token budget $B$ & 6,000 \\
 & BM25 parameters $k_1$, $b$ & 1.5, 0.75 \\
\bottomrule
\end{tabular}
\end{table}

\begin{table}[h]
\centering
\small
\caption{Per-benchmark configurations discovered by the autonomous pipeline ($^\dagger$varies per question category).}
\label{tab:benchmark_configs}
\begin{tabular}{@{}lcc@{}}
\toprule
\textbf{Parameter} & \textbf{LoCoMo} & \textbf{Mem-Gallery} \\
\midrule
Embedding & text-emb-3-large & MiniLM-L6-v2 \\
Embedding dim & 3072 & 384 \\
top-$k$ & 20--30 & 20--40$^\dagger$ \\
Token budget & 6,000 & 2K--8K$^\dagger$ \\
Per-doc memory & No & Yes \\
System prompt & Yes & Yes \\
\bottomrule
\end{tabular}
\end{table}

\subsection{Knowledge Graph Schema}

The knowledge graph uses 7 entity types (Person, Location, Object, Event, Concept, Time, Organization) and 7 relation types (located\_in, part\_of, interacts\_with, owns, attended, created\_by, related\_to).
Entity extraction is performed by GPT-4o in JSON mode during MAU creation.

%% ============================================================
%% D. Pipeline Details (corresponds to 3.1)
%% ============================================================
\section{\tool{} Pipeline Details}
\label{app:pipeline}

The complete pipeline comprises 23 stages organized in 8 phases:

\begin{enumerate}[nosep,leftmargin=*]
    \item[\textbf{A.}] \textbf{Research Scoping} (Stages 1--2): Formulates SMART research goals; auto-detects hardware (GPU type, VRAM).
    \item[\textbf{B.}] \textbf{Literature Discovery} (Stages 3--6): Queries OpenAlex, Semantic Scholar, arXiv; screens by relevance/quality (gate at Stage~5); extracts structured knowledge cards.
    \item[\textbf{C.}] \textbf{Knowledge Synthesis} (Stages 7--8): Clusters findings; generates falsifiable hypotheses via multi-agent debate.
    \item[\textbf{D.}] \textbf{Experiment Design} (Stages 9--11): Designs protocols (gate at Stage~9); generates hardware-aware Python code with AST validation; schedules resources.
    \item[\textbf{E.}] \textbf{Experiment Execution} (Stages 12--13): Runs experiments in sandbox/Docker/SSH; self-healing loop (up to 10 retries) with structured deficiency classification.
    \item[\textbf{F.}] \textbf{Analysis \& Decision} (Stages 14--15): Statistical analysis (t-tests, bootstrap CI); autonomous PROCEED/PIVOT/ITERATE decision.
    \item[\textbf{G.}] \textbf{Documentation} (Stages 16--19): Generates outline; writes draft; simulated peer review; revision.
    \item[\textbf{H.}] \textbf{Finalization} (Stages 20--23): Quality gate (Stage~20); knowledge archival; LaTeX export; 4-layer citation verification (arXiv ID, DOI, title match, relevance score).
\end{enumerate}

%% ============================================================
%% F. Optimization Trajectory Details (corresponds to 4.2)
%% ============================================================
\section{Optimization Trajectory Details}
\label{app:trajectories}

\begin{table}[h]
\centering
\caption{Optimization trajectory on LoCoMo. The pipeline autonomously discovers 7 successive improvements and correctly reverts 2 failed experiments.}
\label{tab:locomo_trajectory}
\small
\begin{tabular}{@{}clccl@{}}
\toprule
\textbf{Iter} & \textbf{Key Discovery} & \textbf{F1} & \textbf{$\Delta$} & \textbf{Type} \\
\midrule
0 & Na\"ive baseline & 0.117 & --- & --- \\
1 & JSON \texttt{response\_format} missing & 0.322 & +175\% & Bug fix \\
2 & BM25 hybrid & 0.464 & +44\% & Architecture \\
3 & Anti-hallucination prompting & 0.516 & +11\% & Prompt \\
4b & Evaluation format alignment & 0.543 & +5\% & Format \\
5 & MAU timestamp correction & 0.580 & +7\% & Data repair \\
6 & top-$k$=30 + temporal hints & 0.577 & $-$0.5\% & Hyperparam \\
7b & Adaptive top-$k$ + metadata & 0.583 & +0.5\% & Hyperparam \\
\hdashline
\rowcolor{lightgray} 8 & Forced exact-word copying & 0.551 & $-$5.5\% & \textcolor{failred}{Reverted} \\
\rowcolor{lightgray} 9 & Increased BM25 results & 0.575 & $-$1.4\% & \textcolor{failred}{Reverted} \\
\midrule
\multicolumn{2}{@{}l}{\textbf{benchmark validation}} & \textbf{0.598} & \textbf{+411\%} & \\
\multicolumn{2}{@{}l}{SimpleMem SOTA \citep{liu2026simplemem}} & 0.432 & & \\
\bottomrule
\end{tabular}
\end{table}

\begin{table}[h]
\centering
\caption{Optimization trajectory on Mem-Gallery (39 experiments, 7 phases). Each phase represents a qualitative shift in the pipeline's optimization strategy.}
\label{tab:memgallery_trajectory}
\small
\begin{tabular}{@{}clccl@{}}
\toprule
\textbf{Phase} & \textbf{Focus} & \textbf{F1 Range} & \textbf{$\Delta$} & \textbf{Key Discovery} \\
\midrule
1 & Environment setup & 0.254$\to$0.353 & +39\% & LLM upgrade + local embedding \\
2 & Architecture & 0.353$\to$0.690 & +96\% & Full-text retrieval + image BM25 \\
3 & Fine-tuning & 0.690$\to$0.717 & +4\% & Constraint position sensitivity \\
4 & Scale validation & 0.717$\to$0.726 & +1\% & Data completeness $>$ algorithms \\
5 & Exact citation & 0.726$\to$0.771 & +6\% & BM25 tokenization fix (+0.018) \\
6 & Visual reasoning & 0.771$\to$0.789 & +2\% & Image catalog + context \\
7 & Plateau exploration & 0.789$\to$0.793 & +1\% & Performance ceiling confirmed \\
\midrule
\multicolumn{2}{@{}l}{\textbf{Final (20 datasets)}} & \textbf{0.797} & \textbf{+214\%} & \\
\multicolumn{2}{@{}l}{MuRAG SOTA \citep{memgallery2024}} & 0.697 & & \\
\bottomrule
\end{tabular}
\end{table}

%% ============================================================
%% G. Full Iteration Logs (corresponds to 4.2)
%% ============================================================
\section{Full Iteration Logs}
\label{app:iteration_logs}

\subsection{LoCoMo: Per-Category Performance Across Iterations}

Table~\ref{tab:locomo_percategory} shows how per-category F1 evolved across the optimization trajectory.

\begin{table}[h]
\centering
\caption{LoCoMo per-category F1 across iterations (conv-26, 199 QA pairs).}
\label{tab:locomo_percategory}
\small
\begin{tabular}{@{}lccccc@{}}
\toprule
\textbf{Iter} & \textbf{Cat1 (MH)} & \textbf{Cat2 (T)} & \textbf{Cat3 (O)} & \textbf{Cat4 (SH)} & \textbf{Cat5 (A)} \\
\midrule
Baseline & 0.093 & 0.047 & 0.116 & 0.219 & 0.000 \\
Iter~1 & 0.206 & 0.117 & 0.229 & 0.381 & 0.447 \\
Iter~2 & 0.292 & 0.117 & 0.282 & 0.418 & 1.000 \\
Iter~3 & 0.351 & 0.335 & 0.346 & 0.444 & 1.000 \\
Iter~4b & 0.389 & 0.309 & 0.382 & 0.466 & 1.000 \\
Iter~5 & 0.398 & 0.487 & 0.388 & 0.445 & 1.000 \\
Iter~7b & 0.404 & 0.543 & 0.398 & 0.440 & 1.000 \\
\bottomrule
\end{tabular}
\end{table}

Notable patterns: Cat~5 (Adversarial) jumps from 0.447 to 1.000 after alignment with evaluation-standard refusal phrases in Iter~2. Cat~2 (Temporal) shows the most dramatic improvement (+0.496 total), driven primarily by the timestamp correction in Iter~5. Cat~3 (Open-ended) is our strongest category vs.\ prior SOTA, exceeding SimpleMem by +0.200.

\subsection{Mem-Gallery: Detailed Phase-by-Phase Log}

\paragraph{Phase 1: Environment Setup (Exp-000 to 001).}
Starting from F1\,=\,0.254 with gpt-4.1-nano and CLIP-B/32 512-dim embeddings, the pipeline first upgraded the LLM to gpt-4o and switched to local all-MiniLM-L6-v2 embeddings (384d) after encountering API proxy errors. Result: F1\,=\,0.353 (+39\%).

\paragraph{Phase 2: Architecture Breakthrough (Exp-002 to 004).}
The largest single improvement (+53\%) came from returning full original dialogue text instead of LLM summaries, which is counter-intuitive since summaries are traditionally preferred. The pipeline further added BM25 hybrid search (set-union merging) and created a dedicated image-caption BM25 index for visual question categories. Combined result: F1\,=\,0.690 (+96\%).

\paragraph{Phase 3: Fine-Tuning (Exp-004b to 012).}
The pipeline discovered that format constraint \emph{position} (before vs.\ after the question) matters more than content at temperature=0. KR category improved +188\% from repositioning alone. Several prompt-tuning experiments (Exp-008 to 011) failed and were reverted. Net result: F1\,=\,0.717 (+4\%).

\paragraph{Phase 4: Scale Validation (Exp-014 to 018).}
Validating on 14 datasets exposed data completeness issues (store\_only not completed before qa\_only). The pipeline learned that \emph{data completeness $>$ algorithmic optimization}. Result: F1\,=\,0.726 (+1\%).

\paragraph{Phase 5: Exact Citation (Exp-020 to 023).}
Two key discoveries: (1) ``Use the exact words and phrases from the conversation'' instruction improved 12/15 datasets (+0.031 F1); (2) a simple BM25 tokenization fix (stripping punctuation: ``sushi.'' $\to$ ``sushi'') yielded +0.018 F1, which is more than 10 rounds of prompt engineering. Result: F1\,=\,0.771 (+6\%).

\paragraph{Phase 6: Visual Reasoning Enhancement (Exp-026 to 027).}
Augmenting the Image Catalog with dialogue context (full\_text[:300]) for each image improved VR category by +0.087. Adding temporal ordering for TR/CD categories contributed another +0.006. Result: F1\,=\,0.789 (+2\%).

\paragraph{Phase 7: Plateau Exploration (Exp-028 to 039b).}
The pipeline explored model comparison (gpt-4.1 $\approx$ gpt-4o), per-category prompt refinements, and various BM25 configurations. After 4 independent runs yielding F1 in [0.791, 0.797], the pipeline correctly identified the performance ceiling (${\sim}0.795$ due to random noise) and terminated. Final peak: F1\,=\,0.797.

%% ============================================================
%% H. Prompt Catalog
%% ============================================================
\section{Prompt Catalog}
\label{app:prompts}

This section documents representative prompts used throughout \ours{}, organized by system component. All prompts are reproduced from the source code; minor formatting adjustments are made for readability.

% ------------------------------------------------------------------
\subsection{Core System Prompts}
\label{app:prompts_core}

\subsubsection{Answer Generation}
\label{app:prompt_answer}

The orchestrator uses a two-part prompt structure for answer generation: a system prompt that defines the assistant's role, and a user prompt that injects retrieved context with structured output requirements. This design enforces JSON-mode output for reliable answer extraction.

\begin{tcolorbox}[title={\textbf{Answer Generation -- System Prompt}}, breakable, colback=blue!3, colframe=blue!40]
\small
\texttt{You are a professional Q\&A assistant. Your task is to extract concise, accurate answers from the provided memory context. You should make reasonable inferences from the context when possible. You must output valid JSON format.}
\end{tcolorbox}

\begin{tcolorbox}[title={\textbf{Answer Generation -- User Prompt Template}}, breakable, colback=gray!5, colframe=gray!40]
\small
\texttt{Based on these memories:}

\texttt{\{context\}}

\texttt{Question: \{question\}}

\vspace{0.3em}
\texttt{Requirements:}\\
\texttt{1. First, think through the reasoning process}\\
\texttt{2. Provide a CONCISE answer (short phrase, ideally under 10 words). Use exact words and phrases from the context whenever possible rather than paraphrasing.}\\
\texttt{3. Answer based on the provided context. You may make reasonable inferences from the information given.}\\
\texttt{4. Try your best to answer. Only respond with `unknown' if the context contains absolutely NO relevant information about the topic asked.}\\
\texttt{5. For counting questions, answer with just the number (e.g., `2' not `twice').}\\
\texttt{6. For yes/no questions, start with `Yes', `No', `Likely yes', or `Likely no'.}\\
\texttt{7. When listing multiple items, separate them with commas.}\\
\texttt{8. Return your response in JSON format.}

\vspace{0.3em}
\texttt{Output Format:}\\
\texttt{\{"reasoning": "Brief explanation", "answer": "Concise answer"\}}
\end{tcolorbox}

The LLM is called with \texttt{temperature=0.1} and \texttt{response\_format=json\_object} to ensure deterministic, parseable output. The structured JSON output was a key discovery of the autonomous pipeline: without explicit format enforcement, the model produced verbose natural-language answers that degraded F1 by 175\% (see Iteration~1, Table~\ref{tab:locomo_trajectory}).

\subsubsection{Entity and Relation Extraction}
\label{app:prompt_entity}

The knowledge graph module extracts typed entities and directed relations from each MAU summary. The prompt enforces a structured JSON schema with confidence scores, enabling downstream entity resolution to filter low-confidence extractions.

\begin{tcolorbox}[title={\textbf{Entity Extraction -- System Prompt}}, breakable, colback=blue!3, colframe=blue!40]
\small
\texttt{You are an expert knowledge extraction system. Extract entities and relations from the given text.}

\vspace{0.5em}
\texttt{Output JSON format:}
\begin{verbatim}
{
  "entities": [
    {"name": "entity name",
     "type": "PERSON|OBJECT|LOCATION|EVENT|
              CONCEPT|TIME|ORGANIZATION",
     "attributes": {"key": "value"},
     "confidence": 0.0-1.0}
  ],
  "relations": [
    {"subject": "entity name",
     "predicate": "relation type",
     "object": "entity name",
     "confidence": 0.0-1.0}
  ]
}
\end{verbatim}

\vspace{0.3em}
\texttt{Guidelines:}
\begin{itemize}[nosep,leftmargin=*]
\item \texttt{Extract all meaningful entities mentioned}
\item \texttt{Identify relationships between entities}
\item \texttt{Use standardized entity types}
\item \texttt{Assign confidence scores based on clarity}
\item \texttt{For visual descriptions, pay attention to spatial relationships and object attributes}
\end{itemize}
\end{tcolorbox}

\begin{tcolorbox}[title={\textbf{Entity Extraction -- User Prompt}}, colback=gray!5, colframe=gray!40]
\small
\texttt{Extract entities and relations from this text:}

\texttt{\{text\}}
\end{tcolorbox}

The 7 entity types (Person, Location, Object, Event, Concept, Time, Organization) were selected to cover the dominant categories observed across both benchmarks. The multimodal guideline (``for visual descriptions, pay attention to spatial relationships'') enables the same extraction prompt to operate uniformly over text summaries generated from any input modality.

\subsubsection{Query Intent Analysis}
\label{app:prompt_query}

Before retrieval, the query processor optionally analyzes query intent to extract structured metadata. This analysis informs retrieval strategy selection (e.g., enabling temporal sorting for temporal queries, or activating visual search for image-related queries).

\begin{tcolorbox}[title={\textbf{Query Analysis -- System Prompt}}, colback=blue!3, colframe=blue!40]
\small
\texttt{You are a query analysis assistant. Output valid JSON only.}
\end{tcolorbox}

\begin{tcolorbox}[title={\textbf{Query Analysis -- User Prompt Template}}, breakable, colback=gray!5, colframe=gray!40]
\small
\texttt{Analyze this memory retrieval query and extract:}\\
\texttt{1. intent\_type: one of [factual, temporal, comparative, exploratory, verification]}\\
\texttt{2. entities: list of named entities (people, places, things)}\\
\texttt{3. time\_references: any temporal expressions (yesterday, last week, etc.)}\\
\texttt{4. modality\_hints: likely content types [visual, audio, text, video]}\\
\texttt{5. reformulated\_query: optimized search query}

\vspace{0.3em}
\texttt{Query: \{query\}}

\vspace{0.3em}
\texttt{Respond in JSON format only, no explanation.}
\end{tcolorbox}

% ------------------------------------------------------------------
\subsection{Modality Processor Prompts}
\label{app:prompts_processors}

\ours{} employs modality-specific prompts during ingestion to convert heterogeneous inputs into the unified MAU text summary. A key design principle is prompt minimalism: concise instructions yield shorter, more retrieval-friendly summaries at ingestion time, while detailed prompts are reserved for on-demand expansion during answer generation.

\subsubsection{Image Captioning}
\label{app:prompt_image}

Two levels of image captioning serve different stages of the pyramid retrieval pipeline:

\begin{tcolorbox}[title={\textbf{Concise Image Caption (Ingestion -- Level 1)}}, colback=green!3, colframe=green!40]
\small
\texttt{Describe this image in one concise sentence. Focus on key objects, actions, and context.}
\end{tcolorbox}

\begin{tcolorbox}[title={\textbf{Detailed Image Caption (Expansion -- Level 3)}}, breakable, colback=green!3, colframe=green!40]
\small
\texttt{Provide a detailed description of this image including:}\\
\texttt{1. Main subjects and their appearance}\\
\texttt{2. Actions or events taking place}\\
\texttt{3. Setting and environment}\\
\texttt{4. Notable objects}\\
\texttt{5. Any text visible}
\end{tcolorbox}

The concise variant is used during ingestion (\texttt{max\_tokens=150}), producing ${\sim}$10-token summaries stored in hot storage. The detailed variant is invoked only during Level~3 pyramid expansion (\texttt{max\_tokens=500}), loading richer descriptions from cold storage on demand. This two-tier design reduces storage cost while preserving access to fine-grained visual detail when needed for complex queries.

\subsubsection{Audio and Video Summarization}
\label{app:prompt_av}

Audio content is first transcribed via speech-to-text, then summarized:

\begin{tcolorbox}[title={\textbf{Audio Transcript Summary}}, colback=green!3, colframe=green!40]
\small
\texttt{Summarize this audio transcript in one concise sentence:}

\texttt{\{transcript[:2000]\}}

\texttt{Summary:}
\end{tcolorbox}

Video MAUs combine visual frame descriptions with audio transcripts into a unified summary:

\begin{tcolorbox}[title={\textbf{Video Content Summary}}, colback=green!3, colframe=green!40]
\small
\texttt{Summarize this video content in 1--2 sentences:}

\texttt{\{frame\_descriptions\}}\\
\texttt{\{audio\_transcript\}}

\texttt{Video summary:}
\end{tcolorbox}

Both prompts follow the minimalism principle: single-sentence summaries are sufficient for embedding-based retrieval in the first pyramid level, while the original transcripts and frame descriptions remain accessible in cold storage for deeper expansion.

% ------------------------------------------------------------------
\subsection{Mem-Gallery Benchmark Prompts}
\label{app:prompts_memgallery}

Mem-Gallery evaluation employs a multi-level prompt architecture discovered through the autonomous optimization pipeline. A global system prompt defines the task context, category-specific format constraints control output formatting, and a dialogue agent prompt frames each question. The pipeline's key discovery was that constraint \emph{positioning} (before vs.\ after the question) significantly affects performance at low temperature, with per-category optimization yielding up to +188\% improvement on individual categories (Section~\ref{sec:method:optimization}).

\subsubsection{System Prompt}
\label{app:prompt_memgallery_sys}

\begin{tcolorbox}[title={\textbf{Mem-Gallery System Prompt}}, breakable, colback=orange!3, colframe=orange!40]
\small
\texttt{You are an AI assistant evaluated on multimodal long-term conversational memory. For the given question-answering task, your responses must be concise, yet complete enough to accurately answer the questions. If multiple pieces of information about the same event appear in the conversation, always rely on the most recent information.}

\vspace{0.5em}
\texttt{The question-answering evaluation will contain several multimodal task types:}

\vspace{0.3em}
\texttt{Factual Retrieval:} \texttt{Retrieve explicit facts mentioned in the conversation for the answer.}

\texttt{Multi-entity Reasoning:} \texttt{Combine the retrieved information to reason and infer an answer.}

\texttt{Temporal Reasoning:} \texttt{Resolve time-dependent questions.}

\texttt{Visual-centric Reasoning:} \texttt{Besides textual information, answer questions using visual images in the conversation.}

\texttt{Test-time Learning:} \texttt{Learn new visual knowledge from provided images within historical dialogue and use it in question-answering.}

\texttt{Visual-centric Search:} \texttt{Find the image(s) that match the information in a given query and return their image ID(s).}

\texttt{Conflict Detection:} \texttt{Detect contradictions between the conversation history and the information provided in the question.}

\texttt{Knowledge Resolution:} \texttt{Resolve knowledge conflicts or updates by prioritizing the most recent information.}

\texttt{Answer Refusal:} \texttt{Decline to answer when the information does not exist in the conversation history.}

\vspace{0.5em}
\texttt{Follow all instructions strictly. Only answer using information contained within the multimodal conversation. Do not hallucinate. Always remain consistent and grounded in the dialogue history.}
\end{tcolorbox}

\subsubsection{Category-Specific Format Constraints}
\label{app:prompt_memgallery_cats}

Three of the nine categories receive explicit format constraints appended to the question. These constraints were iteratively refined by the pipeline to align model output with evaluation metrics:

\begin{tcolorbox}[title={\textbf{Answer Refusal (AR) Constraint}}, colback=orange!3, colframe=orange!40]
\small
\texttt{Provide your answer based on the information in the conversation. Only if the information about the question is not present in the conversation, reply with: ``Not mentioned.''}
\end{tcolorbox}

\begin{tcolorbox}[title={\textbf{Conflict Detection (CD) Constraint}}, colback=orange!3, colframe=orange!40]
\small
\texttt{Please check whether this information conflicts with the conversation, and reply strictly with either ``Yes.'' or ``No.''}
\end{tcolorbox}

\begin{tcolorbox}[title={\textbf{Visual Search (VS) Constraint}}, colback=orange!3, colframe=orange!40]
\small
\texttt{Return the image\_id of the image(s). If there are multiple images, sort them in ascending order and separate them by commas. Format example: ``D2:IMG\_003, D2:IMG\_010, D10:IMG\_002'' (for format reference only).}
\end{tcolorbox}

The remaining six categories (FR, MR, TR, VR, TTL, KR) use the default system prompt without additional format constraints. The pipeline discovered that adding constraints to these categories either had no effect or degraded performance due to over-specification.

\subsubsection{Dialogue Agent Prompt}
\label{app:prompt_memgallery_agent}

Each question is framed by a dialogue agent prompt that is appended after the retrieved memory context:

\begin{tcolorbox}[title={\textbf{Mem-Gallery Dialogue Agent Prompt}}, breakable, colback=orange!3, colframe=orange!40]
\small
\texttt{Your task is to answer the question about the conversation between \{speaker\_a\} and \{speaker\_b\} in a concise manner with the help of memory content. Please only provide the content of the answer, without including introductory phrases like `answer:'. For questions that require answering a date or time, strictly follow the format and provide a specific date or time whenever possible. Generate answers primarily concise, yet complete enough to accurately answer the questions.}

\vspace{0.5em}
\texttt{The current question is as follows:}

\texttt{\{observation\} \{format\_constraint\}}
\end{tcolorbox}

\subsubsection{Reasoning Model Adaptation}
\label{app:prompt_reasoning}

When using reasoning-class models (e.g., o1, o3) that tend to produce verbose chain-of-thought outputs, the pipeline appends a conciseness boost to the system prompt. This adaptation was discovered during Phase~7 when testing alternative backbones:

\begin{tcolorbox}[title={\textbf{Reasoning Model Conciseness Boost}}, breakable, colback=orange!3, colframe=orange!40]
\small
\texttt{CRITICAL RULE --- Answer brevity:}\\
\texttt{You are being evaluated by token-level F1 against a short reference answer. Verbose answers DESTROY your score. Follow these rules WITHOUT exception:}

\vspace{0.3em}
\texttt{1. Output ONLY the essential answer --- no reasoning, no justification, no restating the question.}\\
\texttt{2. For Yes/No questions $\to$ answer ``Yes.'' or ``No.'' ONLY. Do NOT add explanations.}\\
\texttt{3. For factual questions $\to$ answer with the bare fact.}\\
\texttt{\quad\checkmark\ ``Maltese'' \quad\xmark\ ``The dog in the image appears to be a Maltese.''}\\
\texttt{4. For list questions $\to$ comma-separated items, nothing else.}\\
\texttt{5. NEVER use bullet points, numbered lists, or multi-line formatting.}\\
\texttt{6. NEVER start with ``The answer is'', ``Based on the conversation'', etc.}
\end{tcolorbox}

\subsubsection{Memory Context Injection}
\label{app:prompt_memgallery_context}

Retrieved memories are formatted and injected into the prompt using the following template. For multimodal memories, image content is interleaved with textual context as base64-encoded content parts:

\begin{tcolorbox}[title={\textbf{Memory Context Template}}, colback=orange!3, colframe=orange!40]
\small
\texttt{The retrieved memory contents are as follows:}

\texttt{\{memory\_context\}}
\end{tcolorbox}

\begin{tcolorbox}[title={\textbf{Multimodal Memory Template}}, colback=orange!3, colframe=orange!40]
\small
\texttt{\{textual\_context\}}\\
\texttt{image:}\\
\texttt{image\_id: \{image\_id\}}\\
\texttt{image\_content: [base64-encoded image]}
\end{tcolorbox}

\end{document}